\documentclass{article}

\PassOptionsToPackage{numbers, compress}{natbib}
    \usepackage[final]{neurips_2024}

\usepackage[utf8]{inputenc} % allow utf-8 input
\usepackage[T1]{fontenc}    % use 8-bit T1 fonts
\usepackage[colorlinks]{hyperref}       % hyperlinks
\usepackage{url}            % simple URL typesetting
\usepackage{booktabs}       % professional-quality tables
\usepackage{amsfonts}       % blackboard math symbols
\usepackage{nicefrac}       % compact symbols for 1/2, etc.
\usepackage{microtype}      % microtypography
\usepackage{xcolor}         % colors
\usepackage{crossreftools}
\usepackage{graphicx}
\usepackage{subfigure}
\usepackage{amsmath}
\usepackage{amssymb}
\usepackage{mathtools}
\usepackage{amsthm}
\usepackage{mathrsfs}
\usepackage{bbm}
\usepackage{tasks}
\usepackage{wrapfig}
\usepackage[capitalise]{cleveref}
\usepackage{enumitem}
\usepackage{parskip}
\usepackage{pifont}% http://ctan.org/pkg/pifont
\usepackage{makecell}
\usepackage{multicol}
\usepackage{tikz-cd}
\usepackage{units}
\usepackage{titletoc}
\usepackage{appendix}
\usepackage{multirow}

\theoremstyle{plain}
\newtheorem{theorem}{Theorem}[section]
\newtheorem{proposition}[theorem]{Proposition}
\newtheorem{lemma}[theorem]{Lemma}

\theoremstyle{definition}
\newtheorem{definition}[theorem]{Definition}

\theoremstyle{remark}
\newtheorem{remark}[theorem]{Remark}

\providecommand{\calM}{\mathcal{M}}
\providecommand{\calN}{\mathcal{N}}

\providecommand{\calL}{\mathcal{L}}

\providecommand{\bbD}{\mathbb{D}}
\providecommand{\sym}[1]{\mathcal{S}^{#1}}
\providecommand{\spd}[1]{\mathcal{S}^{#1}_{++}}
\providecommand{\cho}[1]{\mathcal{L}_{+}^{#1}}
\providecommand{\tril}[1]{\mathcal{L}^{#1}}
\providecommand{\bbR}[1]{\mathbb {R}^{#1}}
\providecommand{\bbRplus}{\mathbb {R}_{+}}

\providecommand{\orth}[1]{\mathrm{O}({#1})}
\providecommand{\bfst}{\mathbf{ST}}

\providecommand{\rieexp}{\operatorname{Exp}}
\providecommand{\rielog}{\operatorname{Log}}
\providecommand{\diff}{\operatorname{d}}

\providecommand{\pt}[2]{\Gamma_{#1 \rightarrow #2}}

\providecommand{\bbD}{\mathbb {D}}

\providecommand{\diff}{\operatorname{d}}
\providecommand{\tr}{\operatorname{tr}}

\providecommand{\tr}{\operatorname{tr}}
\providecommand{\chol}{\operatorname{Chol}}
\providecommand{\sgn}{\operatorname{sgn}}
\providecommand{\dlog}{\operatorname{Dlog}}
\providecommand{\argmax}{\operatorname{argmax}}
\providecommand{\bzero}{\mathbf{0}}

\providecommand{\st}{\mathrm{s.t.}}
\providecommand{\alphabeta}{(\alpha,\beta)}
\providecommand{\biparamEM}{(\alpha,\beta)\text{-EM}}
\providecommand{\biparamAIM}{(\alpha,\beta)\text{-AIM}}
\providecommand{\biparamLEM}{(\alpha,\beta)\text{-LEM}}
\providecommand{\triparamLEM}{(\theta,\alpha,\beta)\text{-LEM}}
\providecommand{\triparamEM}{(\theta,\alpha,\beta)\text{-EM}}
\providecommand{\triparamAIM}{(\theta,\alpha,\beta)\text{-AIM}}
\providecommand{\paramBWM}{2\theta\text{-BWM}}
\providecommand{\paramLCM}{\theta\text{-LCM}}
\providecommand{\BWM}{\text{BWM}}
\providecommand{\LCM}{\text{LCM}}

\providecommand{\gtriparamEM}{g^{(\theta,\alpha,\beta)\text{-EM}}}

\providecommand{\gparamBWM}{g^{2\theta\text{-BWM}}}

\providecommand{\skewtinize}{\operatorname{skew}}
\providecommand{\so}[1]{\mathrm{SO}(#1)}
\providecommand{\soprod}[2]{\mathrm{SO}^{#1}(#2)}
\providecommand{\soLieAlgebra}[1]{\mathfrak{so}(#1)}
\providecommand{\qr}{\mathcal{Q}}

\providecommand{\graspp}[1]{\mathrm{Gr}(#1)}
\providecommand{\grasonb}[1]{\widetilde{\mathrm{Gr}}(#1)}
\providecommand{\idpp}{I_{p,n}}
\providecommand{\idonb}{\widetilde{I}_{p,n}}

\providecommand{\grassominus}{\ominus_{gr}}
\providecommand{\grassoplus}{\oplus_{gr}}

\providecommand{\spsd}[1]{\mathcal{S}_{#1}^{+}}
\providecommand{\spsdominus}{\ominus_{psd,g}}
\providecommand{\spsdoplus}{\oplus_{psd,g}}

\providecommand{\gominus}{\ominus_{g}}
\providecommand{\goplus}{\oplus_{g}}

\renewcommand{\eqref}[1]{Eq.~(\ref{#1})}

\newcommand{\na}{\textcolor{gray}{N/A}}%
\newcommand{\cmark}{\textcolor{green}{\text{\ding{51}}}}%
\renewcommand{\tiny}{\fontsize{5pt}{6pt}\selectfont}
\providecommand{\ie}{\emph{i.e.,}}
\providecommand{\eg}{\emph{e.g.,}}

\crefname{equation}{Eq.}{Eqs.}
\Crefname{equation}{Equation}{Equations}
\crefname{figure}{Fig.}{Figs.}
\Crefname{figure}{Figure}{Figures}
\crefname{table}{Tab.}{Tabs.}
\Crefname{table}{Table}{Tables}
\crefname{section}{Sec.}{Secs.}
\Crefname{section}{Section}{Sections}
\crefname{appendix}{App.}{Apps.}
\Crefname{appendix}{Appendix}{Appendices}

\crefname{theorem}{Thm.}{Thms.}
\Crefname{theorem}{Theorem}{Theorems}
\crefname{lemma}{Lem.}{Lems.}
\Crefname{lemma}{Lemma}{Lemmas}
\crefname{definition}{Def.}{Defs.}
\Crefname{definition}{Definition}{Definitions}
\crefname{corollary}{Cor.}{Cors.}
\Crefname{corollary}{Corollary}{Corollaries}
\crefname{remark}{Rem.}{Rems.}
\Crefname{remark}{Remark}{Remarks}
\crefname{proposition}{Prop.}{Props.}
\Crefname{proposition}{Proposition}{Propositions}

\input{link_proof}

\title{RMLR: Extending Multinomial Logistic Regression into General Geometries}
\author{Ziheng Chen$^1$, Yue Song$^2$\thanks{Corresponding author}, Rui Wang$^3$, Xiao-Jun Wu$^3$, Nicu Sebe$^1$ \\
$^1$ University of Trento, $^2$ Caltech, $^3$ Jiangnan University\\
\texttt{ziheng\_ch@163.com, yue.song@unitn.it}
}

\begin{document}
\maketitle

\begin{abstract}
Riemannian neural networks, which extend deep learning techniques to Riemannian spaces, have gained significant attention in machine learning. 
To better classify the manifold-valued features, researchers have started extending Euclidean multinomial logistic regression (MLR) into Riemannian manifolds. 
However, existing approaches suffer from limited applicability due to their strong reliance on specific geometric properties.
This paper proposes a framework for designing Riemannian MLR over general geometries, referred to as RMLR.
Our framework only requires minimal geometric properties, thus exhibiting broad applicability and enabling its use with a wide range of geometries. 
Specifically, we showcase our framework on the Symmetric Positive Definite (SPD) manifold and special orthogonal group $\so{n}$, \ie the set of rotation matrices in $\bbR{n}$.
On the SPD manifold, we develop five families of SPD MLRs under five types of power-deformed metrics.
On $\so{n}$, we propose Lie MLR based on the popular bi-invariant metric.
Extensive experiments on different Riemannian backbone networks validate the effectiveness of our framework. The code is available at \url{https://github.com/GitZH-Chen/RMLR}.
\end{abstract}
\section{Introduction}
\label{sec:intro}
In recent years, significant advancements have been achieved in Deep Neural Networks (DNNs), enabling them to effectively analyze complex patterns from various types of data, including images, videos, and speech \citep{hochreiter1997long, krizhevsky2012imagenet,he2016deep,vaswani2017attention}. 
However, most existing models have primarily assumed the underlying data with a Euclidean structure. 
Recently, a growing body of research has emerged, recognizing that the latent spaces of many applications exhibit non-Euclidean geometries, such as Riemannian geometries \citep{bronstein2017geometric}. 
Various frequently-encountered manifolds in machine learning have posed interesting challenges and opportunities, including special orthogonal groups $\so{n}$ \citep{vemulapalli2014human,huang2017deep}, symmetric positive definite (SPD) \citep{huang2017riemannian,brooks2019riemannian,lopez2021vector,wang2024spd,chen2024understanding,chen2024product}, Gaussian \citep{chen2021hybrid,nguyen2021geomnet}, Grassmannian \cite{huang2018building,wang2024grassatt} spherical \citep{skopek2019mixed}, and hyperbolic manifolds~\citep{ganea2018hyperbolic}. 
These manifolds share an important Riemannian property — their Riemannian operators, including geodesics, exponential \& logarithmic maps, and parallel transportation, often possess closed-form expressions. Leveraging these Riemannian operators, researchers have successfully generalized different types of DNNs into manifolds, dubbed \textit{Riemannian neural networks}.

Although Riemannian networks demonstrated success in many applications, most approaches still rely on Euclidean spaces for classification, such as tangent spaces \citep{huang2017riemannian, huang2017deep, brooks2019riemannian,nguyen2021geomnet,wang2021symnet,wang2022learning,nguyen2022gyro,nguyen2022gyrovector,kobler2022spd,wang2022dreamnet,chen2023riemannian}, ambient Euclidean spaces \citep{wang2020deep, song2021approximate, song2022eigenvalues}, or coordinate systems \citep{chakraborty2018statistical}. %Notice that there are also works designing classifiers for graph-based manifold-valued data \citep{chakraborty2020manifoldnet}, but we focus on non-gridded data in line with many previous SPD networks.
However, these strategies distort the intrinsic geometry of the manifold, undermining the effectiveness of Riemannian networks. 
Researchers have recently started directly developing Riemannian Multinomial Logistic Regression (RMLR) on manifolds.
Inspired by the idea of hyperplane margin \citep{lebanon2004hyperplane}, \citet{ganea2018hyperbolic} developed a hyperbolic MLR in the Poincaré ball for Hyperbolic Neural Networks (HNNs).
Motivated by HNNs, \citet{nguyen2023building} developed three kinds of gyro SPD MLRs based on three distinct gyro structures of the SPD manifold.
In parallel, \citet{chen2024spdrmlr} proposed a framework for building SPD MLRs induced by the flat metrics on the SPD manifold. 
\citet{nguyen2024matrix} proposed gyro MLRs for the Symmetric Positive Semi-definite (SPSD) manifold based on the product of gyro spaces.
However, these classifiers often rely on specific Riemannian properties, limiting their generalizability to other geometries. For instance, the hyperbolic MLR \cite{ganea2018hyperbolic} relies on the generalized law of sine, while the gyro MLRs \cite{nguyen2023building,nguyen2024matrix} rely on the gyro structures. 

This paper presents a framework of RMLR over general geometries. In contrast to previous works, our framework only requires the explicit expression of the Riemannian logarithm, which is the minimal requirement in extending the Euclidean MLR into manifolds. 
Since this property is satisfied by many commonly encountered manifolds in machine learning, our framework can be broadly applied to various types of manifolds. Empirically, we showcase our framework on the SPD manifold and rotation matrices.
On the SPD manifold, we systematically propose SPD MLRs under five families of power-deformed metrics. 
We also present a complete theoretical discussion on the geometric properties of these metrics.
In the Lie group of $\so{n}$, we propose Lie MLR based on the widely used bi-invariant metric to build the Lie MLR. 
Our work is the first to extend the Euclidean MLR into Lie groups.
Besides, our framework incorporates several previous Riemannian MLRs, including gyro SPD MLRs in \cite{nguyen2023building}, SPD MLRs in \cite{chen2024spdrmlr}, and gyro SPSD MLRs in \cite{nguyen2024matrix}.

Our SPD MLRs are validated on four SPD backbone networks, including SPDNet \cite{huang2017riemannian} on the radar and human action recognition tasks and TSMNet \cite{kobler2022spd} on the electroencephalography (EEG) classification tasks for the Riemannian feedforward network, RResNet \cite{katsman2023riemannian} on the human action recognition task for the Riemannian residual network, and SPDGCN \cite{zhao2023modeling} on the node classification for the Riemannian graph neural network.
Our Lie MLR is validated on the classic LieNet \cite{huang2017deep} backbone for the human action recognition task.
Compared with previous non-intrinsic classifiers, our MLRs achieve consistent performance gains.
Especially, our SPD MLRs outperform the previous classifiers by \textbf{14.23\%} on SPDNet and \textbf{13.72\%} on RResNet for human action recognition,  and \textbf{4.46\%} on TSMNet for EEG inter-subject classification. 
Furthermore, our Lie MLR can improve both the training stability and performance. 
In summary, our \textbf{main theoretical contributions} are the following:
\textbf{(a)} 
We develop a general framework for designing Riemannian MLR over general geometries, incorporating several previous Riemannian MLRs on different geometries.
\textbf{(b)} We systematically propose 5 families of SPD MLRs based on different geometries of the SPD manifold.
\textbf{(c)} We propose a novel Lie MLR for deep neural networks on $\so{n}$.

% \textbf{Paper structure:} 
% \cref{sec:preliminary} gives a preliminary review of the geometry of the SPD manifold and $\so{n}$. 
% \cref{sec:rmlr} revisits the existing Riemannian MLRs and points out their limitations, then proposes our general framework of Riemannian MLR (\cref{thm:rmlr}).
% % \cref{sec:spd_mlrs} focuses on SPD manifolds, by systematically studying five families of deformed Riemannian metrics (\cref{tab:properties_rie_metrics}), and proposing five families of SPD classifiers induced by these metrics (\cref{tab:spd_mlr}). 
% \cref{sec:lie_mlr} showcase our Riemannian MLR on $\so{n}$, \ie Lie MLR.
% We present the experimental results and some in-depth analysis in \cref{sec:experiments}. Finally, \cref{sec:conclusions} summarizes the conclusions.
% \tbd{aa}

\textbf{Main theoretical results:}
We solve the Riemannian margin distance to the hyperplane in \cref{thm:rie_margin_dist} and present our RMLR framework in \cref{thm:rmlr}. 
As shown in \cref{tab:mlr_as_ours_cases}, our RMLR incorporates several existing MLRs on different geometries. 
\cref{thm:spdmlrs} showcases our RMLR on the SPD manifold under five families of metrics summarized in \cref{tab:properties_rie_metrics}.
To remedy the numerical instability of BWM geometry on the SPD manifold, we also propose a backpropagation-friendly solver for the SPD MLR under BWM in \cref{app:bp_Ly}.
\cref{thm:lie_mlr} proposes the Lie MLR for the Lie group $\so{n}$.
Due to the page limits, we put all the proofs in \cref{app:proof}.

\section{Preliminaries} 
\label{sec:preliminary}

This section provides a brief review of the basic geometries of SPD manifolds and special orthogonal groups. Detailed review and notations are left in \cref{app:preliminaries,app:notations}.

\textbf{SPD manifolds: }
The set of $n \times n$ symmetric positive definite (SPD) matrices is an open submanifold of the Euclidean space $\sym{n}$ of symmetric matrices, referred to as the SPD manifold $\spd{n}$ \cite{arsigny2005fast}.
There are five kinds of popular Riemannian metrics on $\spd{n}$: Affine-Invariant Metric (AIM) \citep{pennec2006riemannian}, Log-Euclidean Metric (LEM) \citep{arsigny2005fast}, Power-Euclidean Metrics (PEM) \citep{dryden2010power}, Log-Cholesky Metric (LCM) \citep{lin2019riemannian}, and Bures-Wasserstein Metric (BWM) \citep{bhatia2019bures}.
Note that, when power equals 1, the PEM is reduced to the Euclidean Metric (EM).
\citet{thanwerdas2023n} generalized AIM, LEM, and EM into two-parameters families of $\orth{n}$-invariant metrics, \ie $(\alpha,\beta)$-AIM, $(\alpha,\beta)$-LEM, and $(\alpha,\beta)$-EM, with $\min (\alpha, \alpha+n \beta)>0$.
We denote the metric tensor of $(\alpha,\beta)$-AIM, $(\alpha,\beta)$-LEM, $(\alpha,\beta)$-EM, LCM, and BWM as $g^{\alphabeta\text{-AIM}}$, $g^{\alphabeta\text{-LEM}}$, $g^{\alphabeta\text{-EM}}$, $g^{\mathrm{LCM}}$, and $g^{\mathrm{BWM}}$, respectively.

\textbf{Rotation matrices: }
The special orthogonal group $\so{n}$ is the set of $n \times n$ orthogonal matrices with unit determinant, the elements of which are also referred to as rotation matrices. As shown in \cite{hall2013lie}, $\so{n}$ forms a Lie group. 
We adopt the widely used bi-invariant Riemannian metric \cite{boumal2011discrete}.

\section{Riemannian multinomial logistic regression}
\label{sec:rmlr}

Inspired by \cite{lebanon2004hyperplane}, \citet{ganea2018hyperbolic,nguyen2023building,chen2024spdrmlr,nguyen2024matrix} extended the Euclidean MLR into hyperbolic, SPD, and SPSD manifolds.
However, these classifiers rely on specific Riemannian properties, such as the generalized law of sines, gyro structures, and flat metrics, which limits their generality.
In this section, we first revisit several existing MLRs and then propose our Riemannian classifiers with minimal geometric requirements. 

\subsection{Revisiting existing multinomial logistic regressions}
\label{subsec:re_exist_MLR}

Given $C$ classes, the Euclidean MLR computes the multinomial probability of each class:
\begin{equation} \label{eq:EMLR_reform_start}
    \forall k \in\{1, \ldots, C\}, \quad p(y=k \mid x) \propto \exp \left(\left\langle a_k, x\right\rangle-b_k\right),
\end{equation}
where $b_k \in \mathbb{R}$, and $x, a_k \in \mathbb{R}^n \backslash \{ \bzero \}$.
As shown in \cite{ganea2018hyperbolic}, the Euclidean MLR can be reformulated by the margin distance to the hyperplane:
\begin{align} 
    \label{eq:EMLR_reformed}
    p(y=k \mid x) \propto &\exp \left(\operatorname{sign}(\langle a_k, x-p_k\rangle)\|a_k\| d (x, H_{a_k, p_k}) \right),\\
    \label{eq:euc_hyperplane}
    H_{a_k, p_k} &=\{x \in \mathbb{R}^n:\langle a_k, x - p_k\rangle=0\},
\end{align}
where $\langle a_k, p_k\rangle =b_k$, and $H_{a_k, p_k}$ is a hyperplane.

\cref{eq:EMLR_reformed,eq:euc_hyperplane} can be naturally extended into manifolds $\calM$ by Riemannian operators: 
\begin{align}
    \label{eq:rmlr_v1}
    p(y=k \mid S) &\propto \exp \left(\operatorname{sign}(\langle \tilde{A}_k, \rielog_{P_k}(S) \rangle_{P_k})\|\tilde{A}_k\|_{P_k} \tilde{d} (S, \tilde{H}_{\tilde{A}_k, P_k}) \right),\\
    \label{eq:r_hyperplane}
    \tilde{H}_{\tilde{A}_k, P_k} &= \{S \in \calM: g_{P_k}( \rielog_{P_k} S, \tilde{A}_k) =0\},
\end{align}
where $P_k \in \calM, \tilde{A}_k \in T_{P_k}\calM \backslash \{ \bzero \}$, $g_{P_k}$ is the Riemannian metric at ${P_k}$, and $\rielog_{P_k}$ is the Riemannian logarithm at ${P_k}$.
The margin distance is defined as an infimum:
\begin{equation} \label{eq:dist_hyperplane}
    \tilde{d} (S, \tilde{H}_{\tilde{A}_k, P_k})) =\inf _{Q \in \tilde{H}_{\tilde{A}_k, P_k}} d(S, Q).
\end{equation}

The MLRs proposed in \cite{lebanon2004hyperplane,ganea2018hyperbolic,nguyen2023building,chen2024spdrmlr} can be viewed as different implementations of \cref{eq:rmlr_v1}-\cref{eq:dist_hyperplane}. To calculate the MLR in \cref{eq:rmlr_v1}, one has to compute the associated Riemannian metrics, logarithmic maps, and margin distance.
The associated Riemannian metrics and logarithmic maps often have closed-form expressions on the frequently-encounter manifolds in machine learning. 
However, the computation of the margin distance can be challenging. 
On the Poincaré ball of hyperbolic manifolds, the generalized law of sines simplifies the calculation of \cref{eq:dist_hyperplane} \citep{ganea2018hyperbolic}. 
However, the generalized law of sines is not universally guaranteed on other manifolds.
Additionally, \citet{chen2024spdrmlr} developed a closed-form solution of margin distance on the SPD manifold under any metric pulled back from Euclidean spaces. For curved manifolds, solving \cref{eq:dist_hyperplane} would become a non-convex optimization problem. 
To address this challenge, \citet{nguyen2023building} defined gyro structures on the SPD manifold and proposed a pseudo-gyrodistance to calculate the margin distance.
Similarly, \citet{nguyen2024matrix} proposed a pseudo-gyrodistance on the SPSD manifold based on the gyro product space.
However, gyro structures do not necessarily exist in general geometries.
\textit{In summary, the aforementioned methods often rely on specific properties of their associated Riemannian metrics, which usually do not generalize to general geometries.}

\subsection{Riemannian multinomial logistic regression} \label{subsec:general_RMLR}
Recalling \cref{eq:rmlr_v1,eq:r_hyperplane}, the least requirement of extending Euclidean MLR into manifolds is the well-definedness of $\rielog_{P_k}(S)$ for each $k$. In this subsection, we will develop Riemannian MLR, which depends solely on the Riemannian logarithm, without additional requirements, such as gyro structures and generalized law of sines. 
In the following, we always assume the well-definedness of the Riemannian logarithm.
We start by reformulating the Euclidean margin distance to the hyperplane from a trigonometry perspective and then present our Riemannian MLR.

As we discussed before, obtaining the margin distance of \cref{eq:dist_hyperplane} could be challenging.
Inspired by \cite{nguyen2023building}, we resort to the perspective of trigonometry to reinterpret Euclidean margin distance.
In Euclidean space, the margin distance is equivalent to
\begin{equation} \label{eq:reform_e_dits_v2}
    d (x, H_{a, p}))=\sin (\angle x p y^*) d(x, p), \quad \text{with } y^*=\underset{y \in H_{a, p} \backslash\{p\}}{\arg \max }(\cos \angle x p y).
\end{equation}
We extend \cref{eq:reform_e_dits_v2} to manifolds by the Riemannian trigonometry and geodesic distance, the counterparts of Euclidean trigonometry and distance.

\begin{definition}[Riemannian Margin Distance] \label{def:riem_margin_dist}
   Let $\tilde{H}_{\tilde{A}, P}$ be a Riemannian hyperplane defined in \cref{eq:r_hyperplane}, and $S \in \calM$. 
    The Riemannian margin distance from $S$ to $\tilde{H}_{\tilde{A}, P}$ is defined as
    \begin{equation} \label{eq:margin_dist_reform_v2}
        d(S,\tilde{H}_{\tilde{A}, P})=\sin (\angle SPY^*) d(S, P),
    \end{equation}
    where $d(S, P)$ is the geodesic distance, and $Y^*=\argmax (\cos \angle SPY)$ with $Y \in \tilde{H}_{\tilde{A}, P} \backslash\{P\}$. 
    The initial velocities of geodesics define $\cos \angle SPY$:
    \begin{equation} \label{eq:riem_trigo}
        \cos \angle SPY=\frac{\langle \rielog_P Y, \rielog_P S \rangle_P}{\|\rielog_P Y\|_P,\|\rielog_P S\|_P},
    \end{equation}
    where  $\langle \cdot,\cdot \rangle_P$ is the Riemannian metric at $P$, and $\| \cdot \|_P$ is the associated norm.
\end{definition}
The Riemannian margin distance in \cref{def:riem_margin_dist} has a closed-form expression.
\begin{theorem} 
    \label{thm:rie_margin_dist} 
    \linktoproof{thm:rie_margin_dist}
    The Riemannian margin distance defined in \cref{def:riem_margin_dist} is given as
    \begin{equation} \label{eq:rie_margin_dist}
         d(S,\tilde{H}_{\tilde{A}, P}) 
        = \frac{|\langle  \rielog_P S, \tilde{A} \rangle _P|}{\| \tilde{A} \|_P}.
    \end{equation}
\end{theorem}
Putting the \cref{eq:rie_margin_dist} into \cref{eq:rmlr_v1}, we can a closed-form expression for Riemannian MLR.

\begin{theorem} [RMLR] 
    \label{thm:rmlr}
    \linktoproof{thm:rmlr}
    Given a Riemannian manifold $\{\calM, g\}$, the Riemannian MLR induced by $g$ is
    \begin{equation} \label{eq:rmlr_final}
        p(y=k \mid S \in \calM) \propto \exp \left( \langle \rielog_{P_k} S,  \tilde{A}_k \rangle_{P_k} \right),
    \end{equation}
    where $P_k \in \calM$, $\tilde{A}_k \in T_{P_k}\calM \backslash \{ \bzero \}$,
    and $\rielog$ is the Riemannian logarithm.
\end{theorem}

$\tilde{A}_k$ in \cref{eq:rmlr_final} can not be directly viewed as a Euclidean parameter, as $\tilde{A}_k \in T_{P_k} \calM$ depends on $P_k$ and $P_k$ varies during the training.
However, the tangent vector $\tilde{A}_k$ can be generated from a tangent space at a fixed point.
Several tricks can be used, such as Riemannian parallel transportation \cite{do1992riemannian}, vector transportation \cite{absil2009optimization}, the differential of Lie group or gyrogroup translation \cite{loring2011introduction,ungar2005analytic}.
Following previous work \cite{ganea2018hyperbolic,chen2024spdrmlr,nguyen2023building}, we focus on parallel transportation and Lie group translation:
\begin{align}
    \label{eq:A_by_pt} 
    \tilde{A}_k &=\Gamma_{Q \rightarrow P_k} A_k,\\
    \label{eq:A_by_lt} 
    \tilde{A}_k &= L_{P_k \odot Q_{\odot}^{-1} *,Q} A_k,
\end{align}
where $Q \in \calM$ is a fixed point, $A_k \in T_Q\calM \backslash \{0\}$, $\Gamma$ is the parallel transportation along geodesic connecting $Q$ and $P_k$, and $L_{P_k \odot Q_{\odot}^{-1} *,Q}$ denotes the differential map at $Q$ of left translation $L_{P_k \odot Q_{\odot}^{-1}}$ with $P_k \odot Q_{\odot}^{-1}$ denoting Lie group product and inverse. 
In this way, $A_k$ lies in a fixed tangent space and, therefore, can be optimized by a Euclidean optimizer.

\begin{remark}
    We make the following remarks w.r.t. our Riemannian MLR.
    
    (a).
    The reformulation of \cref{eq:reform_e_dits_v2} in gyro MLR \cite{nguyen2023building,nguyen2024matrix} and ours are different.
    Gyro MLR adopts gyro trigonometry and gyro distance to reformulate \cref{eq:reform_e_dits_v2}, while our method directly uses Riemannian trigonometry and geodesic distance.   
    
    (b).
    Compared with the MLRs on hyperbolic, SPD, or SPSD manifolds in \cite{ganea2018hyperbolic,nguyen2023building,chen2024spdrmlr,nguyen2024matrix}, our framework enjoys broader applicability, as our framework only requires the Riemannian logarithm. This property is commonly satisfied by most manifolds encountered in machine learning, such as the five metrics on SPD manifolds mentioned in \cref{sec:preliminary}, the invariant metric on $\so{n}$ \cite{boumal2011discrete}, and hyperbolic \& spherical manifolds \citep{cannon1997hyperbolic,skopek2019mixed}. 
    Besides, several existing MLRs on different geometries are special cases of our Riemannian MLR, which are detailed in \cref{tab:mlr_as_ours_cases}.

    (c).
    The well-definedness of the Riemannian logarithm is a much weaker requirement compared to the existence of the gyro structure. 
    The gyro structure not only requires the Riemannian logarithm but also implicitly requires geodesic completeness \citep[Eqs. (1-2)]{nguyen2023building}.
    For instance, on SPD manifolds, EM and BWM \citep{thanwerdas2023n} are incomplete, undermining the well-definedness of gyro operations. 
\end{remark}

\begin{table}[htbp]
\centering
\caption{Several MLRs on different geometries are special cases of our MLR.}
\label{tab:mlr_as_ours_cases}
\resizebox{\linewidth}{!}{
\begin{tabular}{cccc}
\toprule
MLR & Geometries & Requirements & \makecell{Incorporated \\ by Our MLR}\\
\midrule
Euclidean MLR (\cref{eq:EMLR_reform_start}) & Euclidean geometry & \na & \cmark (\cref{app:sec:rmlr_gen_emlr}) \\
\midrule
Gyro SPD MLRs \cite{nguyen2023building} & AIM, LEM \& LCM on $\spd{n}$   & Gyro structures & \cmark (\cref{rmk:extending_spd_mlr})  \\
\midrule
Gyro SPSD MLRs \cite{nguyen2024matrix} & SPSD product gyro spaces & \makecell{Gyro structures} & \cmark (\cref{app:sec:gyro_spsd_mlr_as_rmlr})   \\
\midrule
Flat SPD MLRs \cite{chen2024spdrmlr} & $(\alpha,\beta)$-LEM \& $(\theta)$-LCM on $\spd{n}$ & \makecell{Pullback metrics from \\ the Euclidean space} & \cmark (\cref{rmk:extending_spd_mlr})  \\
\midrule
Ours & General Geometries & Riemannian logarithm & \na \\
\bottomrule
\end{tabular}
}
\end{table}

\section{SPD multinomial logistic regressions}
\label{sec:spd_mlrs}

\begin{figure}[t]
\centering
\includegraphics[width=\linewidth,trim=0 5mm 0 0cm]{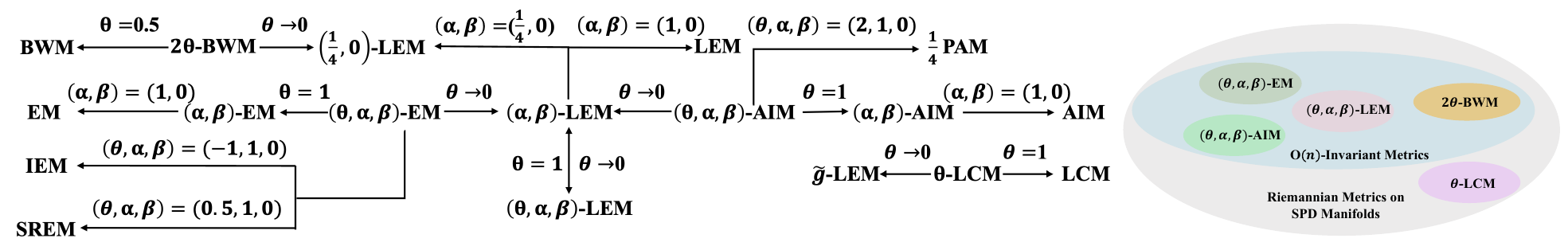}
\caption{Illustration on the deformation (\textbf{left}) and Venn diagram (\textbf{right}) of metrics on SPD manifolds, where IEM, SREM, and $\frac{1}{4}$~PAM denotes Inverse Euclidean Metric, Square Root Euclidean Metric, and Polar Affine Metric scaled by $\nicefrac{1}{4}$.
}
\label{fig:illustration_metrics}
\vspace{-5mm}
\end{figure}

This section showcases our RMLR framework on the SPD manifold.
We first systematically discuss the power-deformed geometries of SPD manifolds.
Based on these metrics, we will develop five families of deformed SPD MLRs.

\subsection{Deformed geometries of SPD manifolds} \label{subsec:geom_spd}

\begin{table}[htbp]
  % \vspace{-2mm}
  \centering
  \caption{Properties of deformed metrics on SPD manifolds ($\theta \neq 0$ and $\min (\alpha, \alpha+n \beta)>0$).}
  \label{tab:properties_rie_metrics}
  \resizebox{0.8\linewidth}{!}{
    \begin{tabular}{cc}
    \toprule
    Name & Properties \\
    \midrule
    $(\theta,\alpha,\beta)$-LEM &
    Bi-Invariance, $\orth{n}$-Invariance, Geodesically Completeness \\  
    \midrule
    $(\theta,\alpha,\beta)$-AIM &
    Lie Group Left-Invariance, $\orth{n}$-Invariance, Geodesically Completeness \\
    \midrule
    $(\theta,\alpha,\beta)$-EM &
    $\orth{n}$-Invariance \\
    \midrule
    $\theta$-LCM &
    Lie Group Bi-Invariance, Geodesically Completeness\\
    \midrule
    $2\theta$-BWM &
    $\orth{n}$-Invariance \\
    \bottomrule    
    \end{tabular}
    }
    % \vspace{-3mm}
\end{table}

As discussed in \cref{sec:preliminary}, there are five popular Riemannian metrics on SPD manifolds. These metrics can be all extended into power-deformed metrics.
For a metric $g$ on $\spd{n}$, the power-deformed metric is defined as
\begin{equation} \label{eq:pow_deform_meric}
    \tilde{g}_P \left(V,W \right) = \frac{1}{\theta^2} g_{P^\theta} \left( (\phi_\theta)_{*,P} (V), (\phi_\theta)_{*,P} (W)\right), \forall P \in \spd{n}, V,W \in T_P\spd{n},
\end{equation}
where $\phi_\theta(P)=P^\theta$ is the matrix power, and $(\phi_\theta)_{*,P}$ is the differential map.
The deformed metric $\tilde{g}$ can interpolate between a LEM-like metric ($\theta \rightarrow 0$) and $g$ ($\theta=1$) \cite{thanwerdas2022geometry}.
Previous work has extended $(\alpha,\beta)$-AIM, $(\alpha,\beta)$-LEM, LCM, and BWM into power-deformed metrics and $(\alpha, \beta)$-LEM is proven to be invariant under the power deformation \cite{chen2024liebn}.
We denote these metrics as $(\theta,\alpha,\beta)$-AIM \cite{thanwerdas2019affine}, $(\alpha, \beta)$-LEM \cite{chen2024liebn}, $2\theta$-BWM \cite{thanwerdas2022geometry}, and $\theta$-LCM \cite{chen2024liebn}, respectively.
The deformation of these metrics is discussed in \cref{app:lim_cases_deformed_metrics}.
We further define the power-deformed metric of $(\alpha, \beta)$-EM by \cref{eq:pow_deform_meric}, denoted as $(\theta,\alpha,\beta)$-EM.
We have the following for the deformation of $(\theta,\alpha,\beta)$-EM.

\begin{proposition} 
    \label{prop:spd_parametrized_metrics}
    \linktoproof{prop:spd_parametrized_metrics}
   $(\theta,\alpha,\beta)$-EM interpolates between $(\alpha,\beta)$-LEM ($\theta \rightarrow 0$) and $\biparamEM$ ($\theta=1$).
\end{proposition}

So far, all five popular Riemannian metrics on SPD manifolds have been generalized into power-deformed families of metrics.
We summarize their associated properties in \cref{tab:properties_rie_metrics} and present their theoretical relation in \cref{fig:illustration_metrics}. 
We leave technical details in \cref{app:props_deformed_metrics}.

\subsection{Five families of SPD multinomial logistic regressions}

\begin{figure}[t]
\centering
\includegraphics[width=\linewidth,trim=0 2cm 0 0]{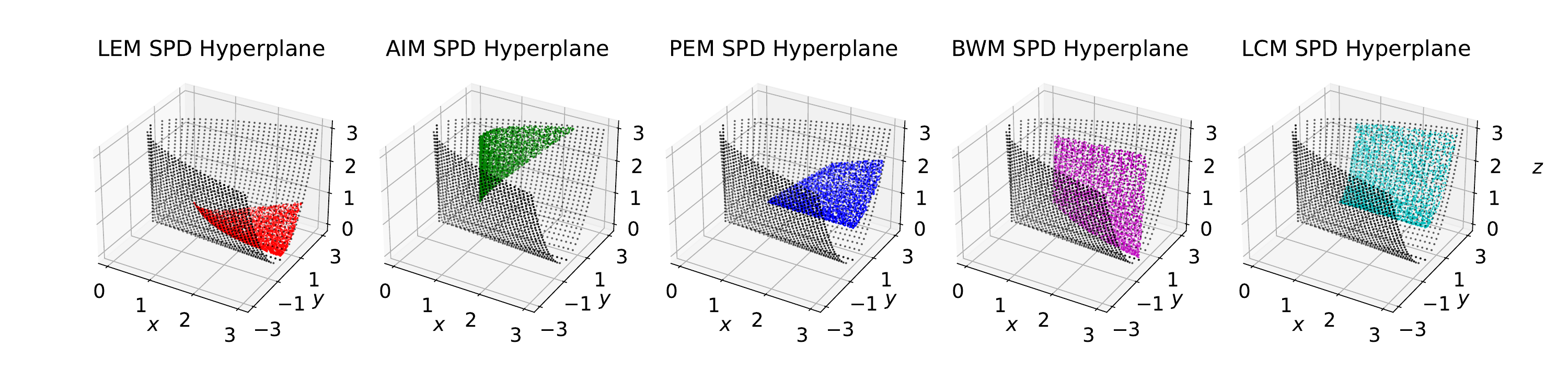}
% trim={left bottom right top}
\caption{Conceptual illustration of SPD hyperplanes induced by five families of Riemannian metrics.
The black dots denote the boundary of $\spd{2}$.
}
\label{fig:hyperplanes}
\vspace{-5mm}
\end{figure}

% \begin{table*}[tbp]
%   \centering
%   %\vspace{-4mm}
%   \caption{Five families of SPD MLRs. For simplicity, we omit the subscripts $k$ of $A_k$ and $P_k$.}
%   \label{tab:spd_mlr}
%   \resizebox{0.8\linewidth}{!}{
%     \begin{tabular}{ccc}
%     \toprule
%     Metrics & $p(y=k \mid S \in \spd{n})=$ & Prototype \\
%     \midrule
%     $\biparamLEM$ &
%     $\exp \left [ \langle \log(S)-\log(P), A \rangle^{\alphabeta} \right ]$ &
%     \cref{eq:rmlr_pm_pt} \\
%     $(\theta,\alpha,\beta)$-AIM &
%     $\exp \left[ \frac{1}{\theta} \langle \log(P^{-\frac{\theta}{2}} S^\theta P^{-\frac{\theta}{2}}), A \rangle^{\alphabeta} \right]$ &
%     \cref{eq:rmlr_pm_pt} \\
%     $(\theta,\alpha,\beta)$-EM &
%     $\exp \left[ \frac{1}{\theta} \langle S^\theta-P^\theta, A \rangle^{\alphabeta} \right]$ &
%     \cref{eq:rmlr_pm_pt} \\
%     $\theta$-LCM &
%     $\exp \left[ \frac{1}{\theta} \langle \lfloor \tilde{K}\rfloor - \lfloor \tilde{L} \rfloor + \left[\dlog(\bbD(\tilde{K}))-\dlog(\bbD(\tilde{L}))\right], \lfloor A \rfloor + \frac{1}{2}\bbD(A) \rangle \right]$ &
%     \cref{eq:rmlr_pm_pt} \\
%     $2\theta$-BWM &
%     $\exp \left[ \frac{1}{4\theta}\langle (P^{2\theta}S^{2\theta})^{\frac{1}{2}} + (S^{2\theta}P^{2\theta})^{\frac{1}{2}} -2P^{2\theta}, \calL_{P^{2\theta}}(\bar{L} A \bar{L}^\top) \rangle \right]$ &
%     \cref{eq:rmlr_pm_lt} \\
%     \bottomrule    
%     \end{tabular}
%     }
%     \vspace{-2mm}
% \end{table*}

This subsection presents five families of specific SPD MLRs by our general framework in \cref{thm:rmlr} and metrics discussed in \cref{subsec:geom_spd}.
We focus on generating $\tilde{A_k}$ by parallel transportation from the identity matrix, except for $2\theta$-BWM.
Since the parallel transportation under $2\theta$-BWM would undermine numerical stability (please refer to \cref{app:num_stab_bwm_mlr} for more details), we resort to a newly developed Lie group operation \cite{thanwerdas2022theoretically}: 
\begin{equation}
    S_1 \odot S_2 = L_1 S_2 L_1^T, \forall S_1, S_2 \in \spd{n}.
\end{equation}
where $L_1=\chol(S_1)$ is the Cholesky decomposition. 

\begin{theorem}[SPD MLRs] 
    \label{thm:spdmlrs} 
    \linktoproof{thm:spdmlrs}
    By abuse of notation, we omit the subscripts $k$ of $A_k$ and $P_k$.
    Given SPD feature $S$, the SPD MLRs, $p(y=k \mid S \in \spd{n})$, are proportional to
    \begin{align}
        \biparamLEM: 
        & \exp \left [ \langle \log(S)-\log(P), A \rangle^{\alphabeta} \right ], \\
        \triparamAIM:
        &\exp \left[ \frac{1}{\theta} \langle \log(P^{-\frac{\theta}{2}} S^\theta P^{-\frac{\theta}{2}}), A \rangle^{\alphabeta} \right], \\
        \triparamEM:
         &\exp \left[ \frac{1}{\theta} \langle S^\theta-P^\theta, A \rangle^{\alphabeta} \right], \\
        \paramLCM: 
        &\exp \left[ \frac{1}{\theta} \langle \lfloor \tilde{K}\rfloor - \lfloor \tilde{L} \rfloor + \left[\dlog(\bbD(\tilde{K}))-\dlog(\bbD(\tilde{L}))\right], \lfloor A \rfloor + \frac{1}{2}\bbD(A) \rangle \right], \\
        \label{eq:bwm_spdmlr}
        \paramBWM:
        & \exp \left[ \frac{1}{4\theta}\langle (P^{2\theta}S^{2\theta})^{\frac{1}{2}} + (S^{2\theta}P^{2\theta})^{\frac{1}{2}} -2P^{2\theta}, \calL_{P^{2\theta}}(\bar{L} A \bar{L}^\top) \rangle \right],
    \end{align}
    where $A \in T_I\spd{n} \backslash \{0\}$ is a symmetric matrix, $\log(\cdot)$ is the matrix logarithm, $\calL_P(V)$ is the solution to the matrix linear system $\calL_P[V] P+ P \calL_P[V]=V$, known as the Lyapunov operator,
    $\dlog(\cdot)$ is the diagonal element-wise logarithm, 
    $\lfloor \cdot \rfloor$ is the strictly lower part of a square matrix, and $\bbD(\cdot)$ is a diagonal matrix with diagonal elements of a square matrix.
    Besides, $\log_{*,P}$ is the differential maps at $P$, $\tilde{K}= \chol(S^\theta)$, $\tilde{L}= \chol(P^\theta)$, and $\bar{L}=\chol(P^{2\theta})$.
\end{theorem}

\begin{wrapfigure}{r}{0.3\textwidth}
    \centering
    \includegraphics[width=0.25\textwidth,trim=0cm 1cm 0cm 0cm]{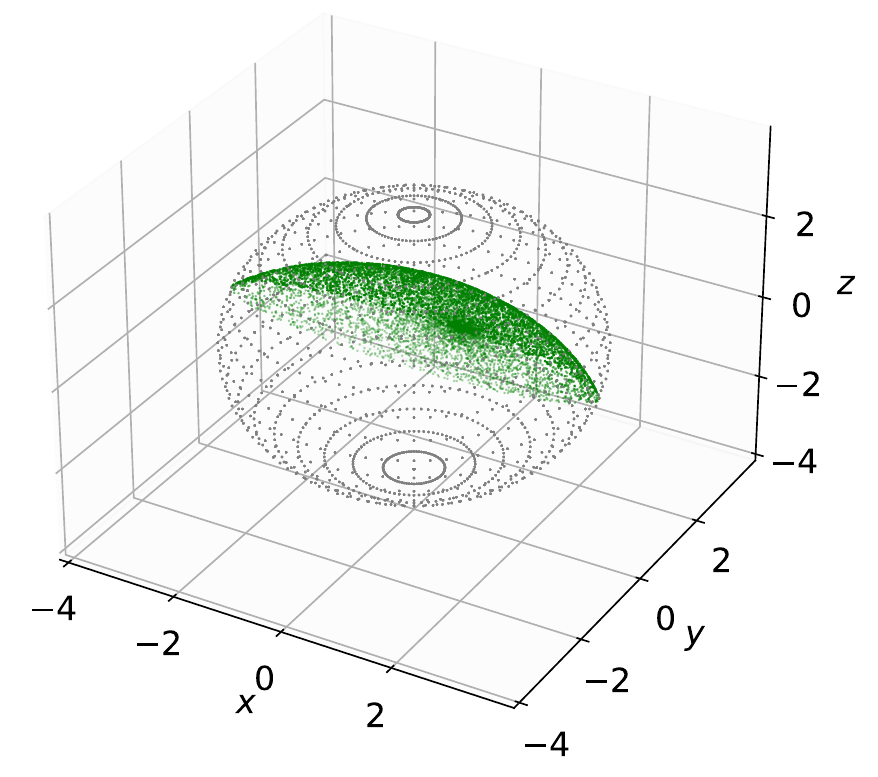}
    % trim={left bottom right top}
    \caption{Conceptual illustration of a Lie hyperplane. Each pair of antipodal black dots corresponds to a rotation matrix with an Euler angle of $\pi$, while the green dots denote a Lie hyperplane.
    }
    \label{fig:rot_hyperplane}
    \vspace{-8mm}
\end{wrapfigure}

The Lyapunov operator in \cref{eq:bwm_spdmlr} requires the eigendecomposition.
However, the backpropagation of eigendecomposition involves $\nicefrac{1}{(\sigma_i-\sigma_j)}$ \cite{ionescu2015training}, undermining the numerical stability.
Therefore, we propose a numerically stable backpropagation for the Lyapunov operator, detailed in \cref{app:bp_Ly}.

As $2 \times 2$ SPD matrices can be embedded into $\bbR{3}$ as an open cone \citep{yair2019parallel}, we illustrate SPD hyperplanes induced by five families of metrics in \cref{fig:hyperplanes}.

\begin{remark} \label{rmk:extending_spd_mlr}
    Our SPD MLRs extend the existing SPD MLRs \cite{nguyen2023building,chen2024spdrmlr}. 
    The pseudo-gyrodistance to a SPD hyperplane in \cite[Thms. 2.23-2.25]{nguyen2023building} is incorporated by our \cref{thm:rie_margin_dist}, while the flat SPD MLRs under $\biparamLEM$ and $\paramLCM$ in \cite[Cor. 4.1]{chen2024spdrmlr} are special cases of our \cref{thm:spdmlrs}. 
    Furthermore, our approach extends the scope of prior work as neither \cite{chen2024spdrmlr} nor \cite{nguyen2023building} explored SPD MLRs based on $\triparamEM$ and $\paramBWM$. 
    The gyro operations in \cite[Eq. (1)]{nguyen2023building} implicitly requires geodesic completeness, whereas $\triparamEM$ and $\paramBWM$ are incomplete. 
    As neither $\triparamEM$ nor $\paramBWM$ belong to pullback Euclidean metrics, the framework presented in \cite{chen2024spdrmlr} cannot be applied to these metrics.
    To the best of our knowledge, our work is the \textbf{first} to apply PEM and BWM to establish Riemannian neural networks, opening up new possibilities for utilizing these metrics in machine learning applications. Besides, neither \citet{nguyen2023building} nor \citet{chen2024spdrmlr} explore the deformed metrics for building SPD MLRs.
\end{remark}

\section{Lie multinomial logistic regression}
\label{sec:lie_mlr}

This section introduces our Lie MLR on $\so{n}$ based on the general RMLR framework in \cref{thm:rmlr}.
The Riemannian metric on $\so{n}$ is assumed to be the invariant metric in \cref{tab:riem_rotation}.

The two ways to generate $\tilde{A}_k$ in RMLR, \ie \cref{eq:A_by_pt,eq:A_by_lt}, are equivalent on $\so{n}$.

\begin{lemma} 
    \label{lem:equi_lie_mlr}
    \linktoproof{lem:equi_lie_mlr}
    \begin{equation}
        \pt{Q}{P} = L_{PQ^{-1} *,Q}, \forall P,Q \in \so{n}.
    \end{equation}
\end{lemma}

Similar with SPD MLRs, we set $Q=I$. 
The Lie MLR on $\so{n}$ is presented in the following.

\begin{theorem} 
    \label{thm:lie_mlr}
    \linktoproof{thm:lie_mlr}
    The Lie MLR on $\so{n}$ is given as
    \begin{equation}
        p(y=k \mid R \in \so{n}) \propto \langle \log(P_k^\top S), A_k \rangle,
    \end{equation}
    where $P_k \in \so{n}$ and $A_k \in \soLieAlgebra{n}$.
\end{theorem}

We refer to the Riemannian hyperplanes (\cref{eq:r_hyperplane}) on $\so{n}$ as Lie hyperplanes.
As $\so{3}$ is homeomorphic to 3-dimensional real projective space $\mathbb{RP}^3$ \cite{hartley2013rotation}, \cref{fig:rot_hyperplane} illustrates Lie hyperplanes in the closed ball in $\bbR{3}$ of radius $\pi$.

\begin{table}[tbp!]
    \centering
    % \vspace{-2mm}
    \caption{Comparison of SPDNet with LogEig against SPD MLRs on the Radar dataset.}
    \label{tb:results_radar}
    \resizebox{0.99\linewidth}{!}{
    \begin{tabular}{c|c|c|cc|cc|cc|cc}
    \toprule
        \multirow{2}[4]{*}{Architectures} & \multirow{2}[4]{*}{LogEig MLR} & $(\theta,\alpha,\beta)$-AIM   & \multicolumn{2}{c|}{$(\theta,\alpha,\beta)$-EM} & \multicolumn{2}{c|}{$(\alpha,\beta)$-LEM} & \multicolumn{2}{c|}{$2\theta$-BWM} & \multicolumn{2}{c}{$\theta$-LCM} \\
    \cmidrule{3-11}          &       & (1,1,0) & (1,1,0) & (1,1,$\nicefrac{1}{8}$) & (1,1,0) & (1,1,1) & (0.5) & (0.25) & (1)   & (0.5) \\
    \midrule
    2-Block & 92.88±1.05 & \textbf{ 94.53±0.95} & 94.24±0.55 & \textbf{94.93±0.60} & 93.55±1.21 & \textbf{95.64±0.83} & 92.22±0.83 & \textbf{94.99±0.47} & 93.49±1.25 & \textbf{94.59±0.8}2 \\
    5-Block & 93.47±0.45 & \textbf{94.32±0.94} & \textbf{95.11±0.82} & 95.01±0.84 & 94.60±0.70 & \textbf{95.87±0.58} & 93.69±0.66 & \textbf{94.84±0.68} & 93.93±0.98 & \textbf{95.16±0.67} \\
    \bottomrule
    \end{tabular}%
    }
    \vspace{-2mm}
\end{table}%

\begin{table}[tbp!]
    \centering
    % \vspace{-2mm}
    \caption{Comparison of SPDNet with LogEig against SPD MLRs on the HDM05 dataset.}
    \label{tb:results_hdm05}
    \resizebox{0.99\linewidth}{!}{
    \begin{tabular}{c|c|c|cc|c|c|cc}
    \toprule
    \multirow{2}[4]{*}{Architectures} & \multirow{2}[4]{*}{LogEig MLR} & $(\theta,\alpha,\beta)$-AIM & \multicolumn{2}{c|}{$(\theta,\alpha,\beta)$-EM} & $(\alpha,\beta)$-LEM & $2\theta$-BWM   & \multicolumn{2}{c}{$\theta$-LCM} \\
    \cmidrule{3-9}          &       & {(1,1,0)} & (1,1,0) & (0.5,1.0,$\nicefrac{1}{30}$) & (1,1,0) & (0.5) & (1) & (0.5) \\
    \midrule
    1-Block & 57.42±1.31 & 58.07±0.64  & 66.32±0.63 & \textbf{71.65±0.88} &  56.97±0.61 & \textbf{70.24±0.92} &  63.84±1.31 &	\textbf{65.66±0.73} \\
    2-Block & 60.69±0.66 &  60.72±0.62 & 66.40±0.87 & \textbf{70.56±0.39} &  60.69±1.02 & \textbf{70.46±0.71} & 62.61±1.46 &	\textbf{65.79±0.63} \\
    3-Block & 60.76±0.80 & 61.14±0.94 & 66.70±1.26 & \textbf{70.22±0.81} &  60.28±0.91 & \textbf{70.20±0.91} &  62.33±2.15 & \textbf{65.71±0.75} \\
    \bottomrule
    \end{tabular}%
    }
    \vspace{-2mm}
\end{table}

\section{Experiments}
\label{sec:experiments}

\begin{table}[tbp!]
    \centering
    % \vspace{-2mm}
    \caption{Inter-session experiments of TSMNet with different MLRs on the Hinss2021 dataset.}
    \label{tb:results_hinss_inter_session}
    \resizebox{0.99\linewidth}{!}{
    \begin{tabular}{c|c|cc|c|c|c|c|c}
    \toprule
    \multirow{2}[4]{*}{Classifiers} & \multirow{2}[4]{*}{LogEig MLR} & \multicolumn{2}{c|}{$(\theta,\alpha,\beta)$-AIM} & $(\theta,\alpha,\beta)$-EM    & $(\alpha,\beta)$-LEM & $2\theta$-BWM   & \multicolumn{2}{c}{$\theta$-LCM} \\
    \cmidrule{3-9}          &       & (1,1,0) & (0.5,1,0.05) & (1,1,0) & (1,1,0) & (0.5) & (1) & (1.5) \\
    \midrule
    Balanced Acc. & 53.83±9.77 &  53.36±9.92 & \textbf{ 55.27±8.68} & \textbf{ 54.48±9.21} &  53.51±10.02 & \textbf{55.54±7.45} &  55.71±8.57 & \textbf{ 56.43±8.79} \\
    \bottomrule
    \end{tabular}
    }
    \vspace{-2mm}
\end{table}%

\begin{table}[tbp!]
    \centering
    % \vspace{-2mm}
    \caption{Inter-subject experiments of TSMNet with different MLRs on the Hinss2021 dataset.}
    \label{tb:results_hinss_inter_subject}
    \resizebox{0.99\linewidth}{!}{
    \begin{tabular}{c|c|cc|cc|c|cc|cc}
    \toprule
    \multirow{2}[4]{*}{Classifiers} & \multirow{2}[4]{*}{LogEig MLR} & \multicolumn{2}{c|}{$(\theta,\alpha,\beta)$-AIM} & \multicolumn{2}{c|}{$(\theta,\alpha,\beta)$-EM} & $(\alpha,\beta)$-LEM & \multicolumn{2}{c|}{$2\theta$-BWM} & \multicolumn{2}{c}{$\theta$-LCM} \\
    \cmidrule{3-11}          &       & (1,1,0) & (1.5,1,0) & (1,1,0) & (1.5,1,$\nicefrac{1}{20}$) & (1,1,0) & (0.5) & (0.75) & (1)   & (0.5) \\
    \midrule
    Balanced Acc. & 49.68±7.88 &  50.65±8.13 & \textbf{ 51.15±7.83} &  50.02±5.81 & \textbf{ 51.38±5.77} &  \textbf{51.41±7.98} &  50.26±7.23 & \textbf{ 51.67±8.73} &  52.93±7.76 & \textbf{ 54.14±8.36} \\
    \bottomrule
    \end{tabular}%
    }
    \vspace{-2mm}
\end{table}%

We first validate our SPD MLRs on four SPD neural networks: SPDNet \cite{huang2017riemannian} and TSMNet \cite{kobler2022spd} for Riemannian feedforward networks, RResNet \cite{katsman2023riemannian} for Riemannian residual networks, and SPDGCN \cite{zhao2023modeling} for Riemannian graph neural networks.
Then, we proceed with experiments of our Lie MLR under the classic LieNet architecture \cite{huang2017deep}.
The classifier in all the above networks is the LogEig MLR (matrix logarithm + FC + softmax), a Euclidean MLR on the tangent space at the identity matrix.
We substitute the original non-intrinsic LogEig MLR in each baseline model with our RMLRs. 
Notably, the gyro SPD MLRs \cite{nguyen2023building} are special cases of our SPD MLRs under the standard AIM, LEM, and LCM ($(\theta,\alpha,\beta)=(1,1,0)$), while flat SPD MLRs \cite{chen2024spdrmlr} are incorporated by our SPD MLRs under $\biparamLEM$ and $\paramLCM$.
More implementation details are presented in \cref{app:sec:exp_details}.

\subsection{Experiments on the proposed SPD MLRs}
In the following, we abbreviate \textit{SPD MLR-metric} as \textit{metric}.
For instance, $\triparamAIM$ denotes the baseline endowed with the SPD MLR induced by $\triparamAIM$ and (1,1,0) as the value of $(\theta,\alpha,\beta)$. 
\subsubsection{Experiments on the Riemannian feedforward network}
We evaluate our SPD MLRs for Riemannian feedforward networks under the SPDNet and TSMNet backbones.
Following \cite{huang2017riemannian,brooks2019riemannian}, on SPDNet, we use the Radar dataset \cite{brooks2019riemannian} for radar recognition and the HDM05 dataset \cite{muller2007documentation} for human action recognition.
TSMNet \cite{kobler2022spd} is one of the state-of-the-art methods for the EEG classification task.
Following \cite{kobler2022spd}, we use the Hinss2021 \cite{hinss_eegdata_2021} dataset. 
For each family of SPD MLRs, we report the SPD MLR induced from the standard metric ($\theta=1,\alpha=1,\beta=0$), and the one induced from the deformed metric with best $(\theta,\alpha,\beta)$. 
Besides, if the standard SPD MLR is already saturated, we only report the results of the standard one.
Under each metric, We highlight the results in bold of our SPD MLR under the best hyperparameters.
We visualize the results in \cref{app:subsec:visulization}.

\textbf{Radar:}
In line with \cite{brooks2019riemannian}, we evaluate our classifiers under two network architectures: 2-Block and 5-Block configurations.
The 10-fold results (mean±std) are presented in \cref{tb:results_radar}.
Note that the SPD MLR induced by standard AIM is saturated.
Generally speaking, our SPD MLRs achieve superior performance against the vanilla LogEig MLR.
Moreover, for most families of metrics, the associated SPD MLRs with proper $(\theta,\alpha,\beta)$ outperform the standard SPD MLR, demonstrating the effectiveness of our parameterization.
Besides, among all SPD MLRs, the ones induced by $\biparamLEM$ achieve the best performance. 

\textbf{HDM05:}
Following \cite{huang2017riemannian}, three architectures are adopted: 1-Block, 2-Block and 3-Block configurations.
The 10-fold results (mean±std) are presented in \cref{tb:results_hdm05}.
Note that the standard SPD MLRs under AIM, LEM, and BWM are already saturated on this dataset.
As the Radar dataset, similar observations can be made on this dataset.
Our SPD MLRs can bring consistent performance gain for SPDNet, and properly selected hyperparameters can bring further improvement.
Particularly, among all the SPD MLRs, the ones based on the $\paramBWM$ and $\triparamEM$ achieve the best performance. 
Compared to the vanilla LogEig MLR, \textbf{the highest performance improvement is 14.23\%}, highlighting our approach's effectiveness.
Notably, since $\paramBWM$ and $\triparamEM$ are geodesically incomplete and not pulled back from a Euclidean space, the SPD MLR under these two metrics can not be derived by the framework of gyro or flat MLR.
This contrast confirms the applicability of our theoretical framework to a broader range of geometries.

\textbf{Hinss2021:}
The results (mean±std) of leaving 5\% out cross-validation are reported in \cref{tb:results_hinss_inter_session,tb:results_hinss_inter_subject}.
Once again, our intrinsic classifiers demonstrate improved performance compared to the LogEig MLR in both inter-session and inter-subject scenarios.
Besides, the SPD MLRs based on $\theta$-LCM achieve the best performance, \textbf{outperforming the vanilla classifier by 2.6\% for inter-session and by 4.46\% for inter-subject}.
This finding highlights the versatility of our framework.

\begin{table}[htbp]
    \centering
    \vspace{-2mm}
    \caption{Comparison of LogEig against SPD MLRs under the RResNet architecture.}
    \label{tab:results_rresnet}
    \resizebox{0.99\linewidth}{!}{
        \begin{tabular}{ccccccc}
        \toprule
         Datasets & LogEigMLR & $\triparamAIM$ & $\triparamEM$ & $\biparamLEM$ & $\paramBWM$ & $\paramLCM$ \\ 
        \midrule
        HDM05 & 58.17 ± 2.07 & 60.23 ± 1.26 & \textbf{71.89 ± 0.60 ($\uparrow$ 13.72)} & 59.44 ± 0.87 & 69.85 ± 0.23 & 65.76 ± 0.96 \\
        NTU60 & 45.22 ± 1.23 & 48.94 ± 0.68 & 52.24 ± 1.25 & 46.99 ± 0.41 & 50.56 ± 0.59 & \textbf{53.63 ± 0.95 ($\uparrow$ 8.41)} \\
        \bottomrule
        \end{tabular}
    }
    \vspace{-2mm}
\end{table}

\subsubsection{Experiments on the Riemannian residual network}
Following \cite{katsman2023riemannian}, we use the HDM05 and NTU60 \cite{shahroudy2016ntu} datasets on the RResNet backbone. 
For the hyperparameter $(\theta,\alpha,\beta)$ in our SPD MLRs, we borrow the best ones from \cref{tb:results_hdm05}.
\cref{tab:results_rresnet} reports the 10-fold and 5-fold results on the HDM05 and NTU datasets.
The SPD MLRs still consistently outperform the vanilla LogEig MLR. Besides, similar to the SPD MLRs under the SPDNet backbone for action recognition (\cref{tb:results_hdm05}), the SPD MLR based on $\paramLCM$, $\paramBWM$, or $\triparamEM$ outperforms the vanilla LogEig MLR by a large margin. 
Especially, \textbf{the highest performance improvement is 13.72\% and 8.4\%} on these two datasets.

\begin{table}[htbp]
\centering
\caption{Comparison of LogEig against SPD MLRs under the SPDGCN architecture.}
\label{tab:results_spdgcn}
\resizebox{0.8\linewidth}{!}{
\begin{tabular}{ccccccc}
    \toprule
    \multirow{2}[0]{*}{Classifiers} & \multicolumn{2}{c}{Disease} & \multicolumn{2}{c}{Cora} & \multicolumn{2}{c}{Pubmed} \\
    \cmidrule{2-7} 
          & Mean±STD & Max   & Mean±STD & Max   & Mean±STD & Max \\
    \midrule
    LogEig MLR & 90.55 ± 4.83 & 96.85 & 78.04 ± 1.27 & 79.6 & 70.99 ± 5.12 & 77.6 \\
    $\triparamAIM$ & 94.84 ± 2.27 & 98.43 & 79.79 ± 1.44 & 81.6 & 77.83 ± 1.08 & \textbf{80} \\
    $\triparamEM$ & 90.87 ± 5.14 & 98.03 & 79.05 ± 1.23 & 81 & 78.16 ± 2.41 & 79.5 \\
    $\biparamLEM$ & \textbf{96.33 ± 2.19} & \textbf{98.82} & \textbf{79.89 ± 0.99} & 81.8 & \textbf{78.16 ± 2.41} & 79.5 \\
    $\paramBWM$ & 91.93 ± 3.64 & 96.85 & 73.46 ± 2.18 & 77.7 & 73.22 ± 4.06 & 78.1 \\
    $\paramLCM$ & 93.01 ± 2.14 & 98.43 & 77.59 ± 1.20 & 80.1 & 74.46 ± 5.81 & 78.9 \\
    \bottomrule
\end{tabular}
}
\end{table}

\subsubsection{Experiments on the Riemannian graph network}
We use SPDGCN \cite{zhao2023modeling} as the backbone network for the Riemannian graph network.
Following \cite{zhao2023modeling}, we use the Disease \cite{anderson1991infectious}, Cora \cite{sen2008collective}, and Pubmed \cite{namata2012query} datasets for node classification.
The 10-fold average and maximum results of the vanilla LogEig MLR against our SPD MLR with best $(\theta,\alpha,\beta)$ are reported in \cref{tab:results_spdgcn}.
Similar to the previous results, our SPD MLRs outperform the LogEig MLR. Besides, the SPD MLR based on $\biparamLEM$ generally achieves the best performance for SPDGCN.

\begin{table}[htbp]
\centering
\caption{Comparison of LogEig against SPD MLRs for direct classification.}
\label{tab:results_direct_classification}
\resizebox{0.99\linewidth}{!}{
\begin{tabular}{ccccc}
    \toprule
    \multirow{2}[0]{*}{Classifiers} & \multirow{2}[0]{*}{Radar} & \multirow{2}[0]{*}{HDM05} & \multicolumn{2}{c}{Hinss2021} \\
          &       &       & Inster-session & Inster-subject \\
    \midrule
    LogEig MLR & 91.93 ± 1.30 & 48.43 ± 1.25 & 39.76 ± 7.60 & 44.66 ± 7.17 \\
    \midrule
    $\triparamAIM$ & 95.21 ± 0.81 & 49.17 ± 1.08 & 41.14 ± 7.26 & 45.89 ± 6.52 \\
    $\triparamEM$ & 92.25 ± 1.20 & 61.60 ± 0.69 & \textbf{45.78 ± 8.51 ($\uparrow$ 6.02)} & 45.84 ± 4.75 \\
    $\biparamLEM$ & 95.09 ± 0.57 & 49.05 ± 0.91 & 40.88 ± 7.46 & \textbf{46.02 ± 5.96 ($\uparrow$ 1.36)} \\
    $\paramBWM$ & 94.89 ± 0.41 & \textbf{66.77 ± 1.34 ($\uparrow$ 18.34)} & 44.84 ± 8.00 & 45.21 ± 7.44 \\
    $\paramLCM$ & \textbf{95.67 ± 0.61 ($\uparrow$ 3.74)} & 58.66 ± 0.51 & 43.17 ± 6.21 & 45.10 ± 6.20 \\
    \bottomrule
\end{tabular}
}
% \vspace{-3mm}
\end{table}
\subsubsection{Ablations of SPD MLRs on direct classification}
For a more straightforward comparison, we compare LogEig against our SPD MLRs for direct classification.
We adopt the Radar, HDM05, and Hinss2021 datasets.
We follow the preprocessing of SPDNet and TSMNet to model features into the SPD manifold and directly use LogEig or our SPD MLRs for classification.
The average results are presented in \cref{tab:results_direct_classification}.
The hyperparameters $(\theta,\alpha,\beta)$ are borrowed from \cref{tb:results_radar,tb:results_hdm05,tb:results_hinss_inter_session,tb:results_hinss_inter_subject}.
Our SPD MLRs consistently outperform the vanilla LogEig MLR.
Particularly on the HDM05 dataset, \textbf{the highest performance improvement by our SPD MLRs is 18.34\%}, surpassing the non-intrinsic LogEig MLR by a large margin. 
Ablations on model efficiency are also discussed in \cref{app:efficiency}.

\subsection{Experiments on the proposed Lie MLR}

\begin{table}[htbp]
    % \vspace{-2mm}
    \centering
    \caption{Results of LogEig MLR against Lie MLR under the LieNet architecture.}
    \begin{tabular}{ccccc}
    \toprule
    \multirow{2}[0]{*}{Classifiers} & \multicolumn{2}{c}{G3D} & \multicolumn{2}{c}{HDM05} \\
    \cmidrule{2-5}          & Mean±STD & Max   & Mean±STD & Max \\
    \midrule
    LogEig MLR & 87.91±0.90 & 89.73 & 76.92±1.27 & 79.11 \\
    Lie MLR & \textbf{89.13±1.7} & \textbf{92.12} & \textbf{78.24±1.03} & \textbf{80.25} \\
    \bottomrule
    \end{tabular}
    \label{tab:results_lie_mlr}%
    % \vspace{-2mm}
\end{table}%

We apply our Lie MLR into the classic $\so{n}$ network, \ie LieNet \cite{huang2017deep}, where features are on the Lie group of $\so{3} \times \cdots \times \so{3}$. 
% We substitute the original non-intrinsic LogEig classifier with our Lie MLR.
% We refer to LieNet with our Lie MLR as LieNet+LieMLR.
Following LieNet \cite{huang2017deep}, we use G3D \cite{bloom2012g3d} and HDM05 \cite{muller2007documentation} datasets.
We also extend the Riemannian optimization package geoopt \cite{becigneul2018riemannian} into $\so{3}$, allowing for Riemannian optimization. 
We find that Riemannian SGD performs best for LieNet.
\cref{tab:results_lie_mlr} presents the 10-fold average results of LieNet with or without Lie MLR. 
Note that on the HDM05 datasets, the LieNet might fail to converge, fluctuating between the validation accuracy of 70\% - 75\%.
Therefore, we select 10-fold best performance out of 20-fold experiments.
It can be observed that our Lie MLR can improve the performance of LieNet.
Besides, our Lie MLR can also improve the training stability.
On the HDM05 dataset, LieNet fails to converge in 8 out of 20 folds.
However, when endowed with our Lie MLR, LieNet+LieMLR only encounters convergence failures in 2 folds.
\section{Conclusions}
\label{sec:conclusions}
This paper presents a novel and versatile framework for designing RMLR for general geometries, with a specific focus on SPD manifolds and $\so{n}$.
On the SPD manifold, we systematically explore five families of Riemannian metrics and utilize them to construct five families of deformed SPD MLRs.
On $\so{n}$, we develop the Lie MLR for classifying rotation matrices.
Extensive experiments demonstrate the superiority of our intrinsic classifiers. 
We expect that our work could present a promising direction for designing intrinsic classifiers on diverse geometries.

\begin{ack}
This work was partly supported by the MUR PNRR project FAIR (PE00000013) funded by the NextGenerationEU, the EU Horizon project ELIAS (No. 101120237), and a donation from Cisco. The authors also gratefully acknowledge the financial support from the China Scholarship Council (CSC).
\end{ack}

\bibliographystyle{plainnat}
\bibliography{ref.bib}

\clearpage
\appendix
\startcontents[appendices]
\printcontents[appendices]{l}{1}{\section*{Appendix Contents}}
\newpage
%---------------------------------------------------------------------
\section{Limitations and future avenues}
\label{app:sec:limitation_future_work}

\textbf{Limitation:}
Recalling our RMLR in \cref{eq:rmlr_final}, our RMLR might be over-parameterized. In our RMLR, each class would require a Riemannian parameter $P_k$ and Euclidean parameter $A_k$.
Consequently, as the number of classes grows, the classification layer would become burdened with excessive parameters.
We will address this problem in future work.

\textbf{Future work:} 
We highlight the advantage of our approach compared to existing methods that our framework only requires the Riemannian logarithm, which is commonly satisfied by various manifolds encountered in machine learning.
Therefore, as a future avenue, our framework offers various possibilities for designing intrinsic classifiers for neural networks on other manifolds.

\section{Preliminaries} \label{app:preliminaries}
\subsection{Notations} \label{app:notations}
We briefly summarize the notations in \cref{tab:sum_notaitons} for better clarity.

\begin{table}[htbp]
    \centering
    \caption{Summary of notations.}
    \label{tab:sum_notaitons}
    \resizebox{\linewidth}{!}{
    \begin{tabular}{cc}
        \toprule
        Notation & Explanation  \\
        \midrule
        $\{\calM , g \}$ or abbreviated as $\calM$ & A Riemannian manifold \\
        $T_P\calM$ & The tangent space at $P \in \calM$\\
        $g_P(\cdot ,\cdot)$ or $\langle \cdot, \cdot \rangle_P$ & The Riemannian metric at $P \in \calM$ \\ 
        $\| \cdot \|_P$ & The norm induced by $\langle \cdot, \cdot \rangle_P$ on $T_P\calM$ \\
        $\rielog_P$ & The Riemannian logarithm at $P$\\
        $\Gamma_{P \rightarrow Q}$ & The Riemannian parallel transportation along the geodesic connecting $P$ and $Q$\\
        $H_{a,p}$ & The Euclidean hyperplane\\
        $\tilde{H}_{\tilde{A}, P}$ & The Riemannian hyperplane\\
        $\odot$ & A Lie group operation \\
        $\{\calM, \odot\}$ & A Lie group\\
        $P^{-1}_{\odot}$ & The group inverse of $P$ under $\odot$ \\
        $L_P$ & The Lie group left translation by $P \in \calM$ \\
        $f_{*,P}$ & The differential map of the smooth map $f$ at $P \in \calM$\\
        $f^*g$ & The pullback metric by $f$ from $g$\\
        $\spd{n}$ & The SPD manifold \\
        $\so{n}$ & The special orthogonal group \\
        $\sym{n}$ & The Euclidean space of symmetric matrices \\
        $\cho{n}$ & The Cholesky manifold, \ie the set of lower triangular matrices with positive diagonal elements \\
        $\tril{n}$ & The Euclidean space of $n \times n$ real lower triangular matrices \\
        $\langle \cdot, \cdot \rangle$ or $\cdot : \cdot$ & The standard Frobenius inner product\\
        $\bfst$ & $\bfst = \{(\alpha,\beta) \in \mathbb{R}^2 \mid \min (\alpha, \alpha+n \beta)>0\}$ \\
        $\langle \cdot, \cdot \rangle^{\alphabeta}$ & The $\orth{n}$-invariant Euclidean inner product \\
        $g^{\biparamLEM}$ & The Riemannian metric of $\biparamLEM$ \\
        $g^{\biparamAIM}$ & The Riemannian metric of $\biparamAIM$ \\
        $g^{\biparamEM}$ & The Riemannian metric of $\biparamEM$ \\
        $g^{\BWM}$ & The Riemannian metric of BWM \\
        $g^{\LCM}$ & The Riemannian metric of LCM \\
        $\log(\cdot)$ & The matrix logarithm \\
        $\chol(\cdot)$ & Cholesky decomposition\\
        $\dlog(\cdot)$ & The diagonal element-wise logarithm \\
        $\lfloor \cdot \rfloor$ & The strictly lower triangular part of a square matrix \\
        $\bbD(\cdot)$  & A diagonal matrix with diagonal elements from a square matrix\\
        $\calL_{P}[\cdot]$ & The Lyapunov operator \\
        $(\cdot)^{\theta}$ or $\phi_{\theta}(\cdot)$ & The matrix power \\
        $\qr(\cdot)$ & Return an orthogonal matrix by QR decomposition \\
        $\skewtinize(\cdot)$ & $\skewtinize(A)=\frac{A-A^\top}{2}$ \\
        \bottomrule
    \end{tabular}
    }
\end{table}

\subsection{Brief review of Riemannian geometry}
Intuitively, manifolds are locally Euclidean spaces.
Differentials are the generalization of derivatives in classic calculus.
For more details on smooth manifolds, please refer to \cite{loring2011introduction,lee2013smooth}.
Riemannian manifolds are the manifolds endowed with Riemannian metrics, which can be intuitively viewed as point-wise inner products.

\begin{definition}[Riemannian Manifolds] \label{def:riem_manifold}
A Riemannian metric on $\calM$ is a smooth symmetric covariant 2-tensor field on $\calM$, which is positive definite at every point.
A Riemannian manifold is a pair $\{\calM,g\}$, where $\calM$ is a smooth manifold and $g$ is a Riemannian metric.
\end{definition}

W.l.o.g., we abbreviate $\{\calM,g\}$ as $\calM$.
The Riemannian metric $g$ induces various Riemannian operators, including the geodesic, exponential, and logarithmic maps, and parallel transportation. 
These operators correspond to straight lines, vector addition, vector subtraction, and parallel displacement in Euclidean spaces, respectively \citep[Tabel 1]{pennec2006riemannian}.
A plethora of discussions on Riemannian geometry can be found in \cite{do1992riemannian}.

When a manifold $\calM$ is endowed with a smooth operation, it is referred to as a Lie group.

\begin{definition}[Lie Groups] \label{def:lie_group}
A manifold is a Lie group, if it forms a group with a group operation $\odot$ such that $m(x,y) \mapsto x \odot y$ and $i(x) \mapsto x_{\odot}^{-1}$ are both smooth, where $x_{\odot}^{-1}$ is the group inverse of $x$.
\end{definition}

Lastly, we review the definition of pullback metric, a common technique in Riemannian geometry.
This idea is a natural generalization of bijection from set theory.
\begin{definition} [Pullback Metrics] 
    Suppose $\calM,\calN$ are smooth manifolds, $g$ is a Riemannian metric on $\calN$, and $f:\calM \rightarrow \calN$ is smooth.
    Then the pullback of $g$ by $f$ is defined point-wisely,
    \begin{equation}
        (f^*g)_p(V_1,V_2) = g_{f(p)}(f_{*,p}(V_1),f_{*,p}(V_2)),
    \end{equation}
    where $p \in \calM$, $f_{*,p}(\cdot)$ is the differential map of $f$ at $p$, and $V_i \in T_p\calM$.
    If $f^*g$ is positive definite, it is a Riemannian metric on $\calM$, which is called the pullback metric defined by $f$.
\end{definition}

\begin{table}[htbp]
    \centering
    \caption{The associated Riemannian operators and properties of five basic metrics on SPD manifolds.}
    \label{tab:riem_operators_props}
    \resizebox{0.99\linewidth}{!}{
    \begin{tabular}{ccccc}
         \toprule
         Metrics & $g_P(V,W)$ & $\rielog_P Q$ & $\Gamma_{P \rightarrow Q} (V)$ & Properties \\
         \midrule
         $\biparamLEM$ & 
         $\langle \log_{*,P} (V), \log_{*,P} (W) \rangle^{\alphabeta}$ &
         $(\log_{*,P})^{-1} \left[ \log(Q) - \log(P) \right]$  & 
         $(\log_{*,Q})^{-1} \circ \log_{*,P} (V)$ & 
         \makecell{$\orth{n}$-Invariance, \\ Geodesically Completeness}\\ 
         \midrule
         $\biparamAIM$ & 
         $\langle P^{-1}V, W P^{-1} \rangle^{\alphabeta}$ &
         $P^{1 / 2} \log \left(P^{-1 / 2} Q P^{-1 / 2}\right) P^{1 / 2}$ & 
         $(Q P^{-1})^{1 / 2} V (P^{-1} Q)^{1 / 2}$ &
         \makecell{Lie Group Left-Invariance, \\ $\orth{n}$-Invariance, \\ Geodesically Completeness}\\
         \midrule
         $\biparamEM$ &
         $\langle V, W \rangle^{\alphabeta}$ &
         $Q-P$ &
         $V$ &
         $\orth{n}$-Invariance\\
         \midrule
         LCM & $\sum_{i>j} \tilde{V}_{i j} \tilde{W}_{i j}+\sum_{j=1}^n \tilde{V}_{j j} \tilde{W}_{j j} L_{j j}^{-2}$ & 
         $ (\chol^{-1})_{*, L} \left[ \lfloor K\rfloor-\lfloor L\rfloor + \bbD(L) \dlog (\bbD(L)^{-1} \bbD(K)) \right]$ & 
         $(\chol^{-1})_{*, K} \left[\lfloor \tilde{V} \rfloor+\bbD(K) \bbD(L)^{-1} \bbD(\tilde{V}) \right]$& 
         \makecell{Lie Group Bi-Invariance, \\ Geodesically Completeness}\\ 
         \midrule
         BWM & 
         $\frac{1}{2} \langle \calL_P[V], W \rangle$ &
         $(P Q)^{1 / 2}+(Q P)^{1 / 2}-2 P$ &
         $U\left[\sqrt{\frac{\delta_i+\delta_j}{\sigma_i+\sigma_j}}\left[U^{\top} V U\right]_{i j}\right] U^{\top}$ &
         $\orth{n}$-Invariance \\
         \bottomrule
    \end{tabular}  
    }
\end{table}

\subsection{Basic geometries of SPD manifolds} 
\label{app:basic_gem_spd}

Let $\spd{n}$ be the set of $n \times n$ symmetric positive definite (SPD) matrices.
As shown in \cite{arsigny2005fast}, $\spd{n}$ is an open submanifold of the Euclidean space $\sym{n}$ of symmetric matrices.
There are five kinds of popular Riemannian metrics on $\spd{n}$: Affine-Invariant Metric (AIM) \citep{pennec2006riemannian}, Log-Euclidean Metric (LEM) \citep{arsigny2005fast}, Power-Euclidean Metrics (PEM) \citep{dryden2010power}, Log-Cholesky Metric (LCM) \citep{lin2019riemannian}, and Bures-Wasserstein Metric (BWM) \citep{bhatia2019bures}.
Note that, when power equals 1, the PEM is reduced to the Euclidean Metric (EM).
The standard LEM, AIM, and EM have been generalized into parametrized families of metrics.
We define $\bfst = \{(\alpha,\beta) \in \mathbb{R}^2 \mid \min (\alpha, \alpha+n \beta)>0\}$, and denote the $\orth{n}$-invariant Euclidean metric on $\sym{n}$ \citep{thanwerdas2023n} as 
\begin{equation} \label{eq:oim_sym}
    \langle V,W \rangle^{(\alpha, \beta)}=\alpha \langle V,W \rangle + \beta \tr(V)\tr(W),
\end{equation}
where $(\alpha,\beta) \in \bfst$, and $\langle \cdot,\cdot \rangle$ is the Frobenius inner product.
By $\orth{n}$-invariant Euclidean metric on $\sym{n}$, \citet{thanwerdas2023n} generalized AIM, LEM, and EM into two-parameters families of $\orth{n}$-invariant metrics, \ie $(\alpha,\beta)$-AIM, $(\alpha,\beta)$-LEM, and $(\alpha,\beta)$-EM, with $(\alpha,\beta) \in \bfst$.
We denote the metric tensor of $(\alpha,\beta)$-AIM, $(\alpha,\beta)$-LEM, $(\alpha,\beta)$-EM, LCM, and BWM as $g^{\alphabeta\text{-AIM}}$, $g^{\alphabeta\text{-LEM}}$, $g^{\alphabeta\text{-EM}}$, $g^{\mathrm{LCM}}$, and $g^{\mathrm{BWM}}$, respectively.

For any SPD points $P, Q \in \spd{n}$ and tangent vectors $V, W \in T_P\spd{n}$, we follow the notations in \cref{tab:sum_notaitons} and further denote $\tilde{V}=\chol_{*,P}(V)$, $\tilde{W}=\chol_{*,P}(W)$, $L=\chol{P}$, and $K=\chol{Q}$.
For parallel transportation under the BWM, we only present the case where $P, Q$ are commuting matrices, \ie $P=U\Sigma U^\top$ and $Q=U \Delta U^\top$.
We summarize the associated Riemannian operators and properties in \cref{tab:riem_operators_props}.
Although there also exist other metrics on SPD manifolds \citep{thanwerdas2019exploration,thanwerdas2022geometry,thanwerdas2023n}, their lack of closed-form Riemannian operators makes them problematic to be applied in machine learning.

\subsection{Basic geometry of rotation matrices}

\begin{table}[htbp]
    \centering
    \caption{The associated Riemannian operators on Rotation matrices.}
    \label{tab:riem_rotation}
    \resizebox{\linewidth}{!}{
    \begin{tabular}{cccccc}
    \toprule
    Operators & $g_R(A_1,A_2)$ & $\rielog_R S$ & $\pt{R}{S}(A)$ & Projection Map $\Pi_{R}(U)$ & Retraction of $A \in T_R\so{n}$ at $R$ \\
    \midrule
    Expression & $\langle A_1, A_2 \rangle$ & $\log(R^\top S)$ & $A$ & $\skewtinize(R^\top U)$ & $Q = \qr(R + RA)$ \\
    \bottomrule
    \end{tabular}
    } 
\end{table}

% The special orthogonal group $\so{n}$ is the set of $n \times n$ orthogonal matrices with unit determinant, the elements of which are also referred to as rotation matrices. As shown in \cite{hall2013lie}, $\so{n}$ forms a Lie group. The widely used bi-invariant Riemannian metric on $\so{n}$ \cite{boumal2011discrete} is defined as
% \begin{equation} \label{eq:metric_son}
%     g_R(A_1,A_2)=\langle A_1, A_2 \rangle,
% \end{equation}
% where $R \in \so{n}$ and $A_1,A_2 \in T_R\so{n}$. Note that there are two equivalent representations for the tangent vector on $\so{n}$. In this paper, we use the Lie algebra representation, \ie $T_R\so{n} \cong \soLieAlgebra{n}$ as the set of skew-symmetric matrices. 

We denote $R \in \so{n}$, $A_1, A_2 \in T_R\so{n}$, $U \in \bbR{n \times n}$, $\skewtinize(A)=\frac{A-A^\top}{2}$, and $\qr(\cdot)$ as the function return an orthogonal matrix by QR decomposition.
There are two equivalent representations for the tangent vector on $\so{n}$. 
In this paper, we use the Lie algebra representation, \ie $T_R\so{n} \cong \soLieAlgebra{n}$ as the set of skew-symmetric matrices. 
We summarize all the necessary Riemannian ingredients for $\so{n}$ in \cref{tab:riem_rotation}.

For the specific case of $R \in \so{3}$, $R$ can be represented by the Euler angle and axis \citep[Sec. 3.2]{hartley2013rotation}:
\begin{align}
    \theta(R) 
    &= \arccos \left(\frac{\tr( R )-1}{2}\right),\\
     {\omega}\left( {R} \right) 
     &=\frac{1}{2 \sin \left(\theta\left( {R} \right)\right)}
    \left(\begin{array}{l}
     {R} (3,2)- {R} (2,3) \\
     {R} (1,3)- {R} (3,1) \\
     {R} (2,1)- {R} (1,2)
    \end{array}\right).
\end{align}

Besides, the matrix logarithm on $\so{3}$ can be calculated without decomposition \citep[Ex. A.14]{murray2017mathematical}:
\begin{equation}
    \log (R)= \begin{cases}0, & \text { if } \theta( {R})=0 \\ \frac{\theta( {R})}{2 \sin (\theta( {R}))}\left( {R}- {R}^T\right), & \text { otherwise }\end{cases},
\end{equation}
where $\theta$ is the Euler angle. Obviously, the matrix logarithm is related to the Euler angle and axis when $\theta \neq 0$.

\section{RMLR as a natural extension of the Euclidean MLR}
\label{app:sec:rmlr_gen_emlr}
\begin{proposition}
    When $\calM=\bbR{n}$ is the standard Euclidean space, the RMLR defined in \cref{thm:rmlr} becomes the Euclidean MLR in \cref{eq:EMLR_reform_start}.
\end{proposition}
\begin{proof}
    On the standard Euclidean space $\bbR{n}$, $\rielog_y x=x-y, \forall x,y \in \bbR{n}$.
    Besides, the differential maps of left translation and parallel transportation are the identity maps.
    Therefore, given $x,p_k \in \bbR{n}$ and $a_k \in \bbR{n} / \{0\} \cong T_0\bbR{n} / \{0\}$, we have
    \begin{align}
        p(y=k \mid x \in \bbR{n}) 
        &\propto \exp ( \langle \rielog_{p_k} x,  a_k \rangle_{p_k}), \\
        &\propto \exp ( \langle x - p_k,  a_k \rangle),\\
        &\propto \exp ( \langle x,  a_k \rangle -b_k),
    \end{align}
    where $b_k=\langle x,  p_k \rangle$.
\end{proof}

\section{Gyro SPSD MLR as special cases of our RMLR}
\label{app:sec:gyro_spsd_mlr_as_rmlr}

Gyro SPSD MLR \cite{nguyen2024matrix} is derived by the product of the Grassmannian and SPD gyro spaces.
This section will show that the gyro SPSD MLR is the special case of our RMLR on the product geometry of the SPSD manifold. We first review some necessary results about gyro SPSD MLR and then show the equivalence.

Following the notations in \cite{nguyen2024matrix}, we denote the Grassmannian with canonical metric under the projector and ONB perspective as $\graspp{p,n}$ and $\grasonb{p,n}$, respectively.
The space of $n \times n$ SPSD matrices with a fixed rank $p$, denoted as $\spsd{n,p}$, forms an SPSD manifold \citep{bonnabel2013rank}.
As shown in \cite{bonnabel2013rank,nguyen2024matrix}, the SPSD manifold is a product space, \ie $\spsd{n,p} \cong \grasonb{p,n} \times \spd{p}$.
In other words, every $P \in \spsd{n,p}$ can be decomposed as $P=U_P S_P U_P^\top$ with $U_p \in \grasonb{p,n}$ and $S_P \in \spd{p}$.
We further denote $\spd{p,g}$ as the SPD manifold with metric $g$, where $g$ could be AIM, LEM, and LCM.
As shown in \cite{nguyen2024matrix}, the gyro space in $\spsd{n,p}$ can be defined by the product of gyro spaces of $\grasonb{p,n}$ and $\spd{p,g}$.
By this product structure, \citet{nguyen2024matrix} proposed the SPSD Pseudo-gyrodistance to a hyperplane.

\begin{definition}(SPSD Hypergyroplanes \cite{nguyen2024matrix})
    \label{app:def:spsd_hyperplane}
    Let $P, W \in \grasonb{p,n} \times \spd{n,g}$. Then hypergyroplanes in structure space $\grasonb{p,n} \times \spd{n,g}$ are defined as
    \begin{equation}
        H_{W, P}^{p s d, g}=\left\{Q \in \grasonb{p,n} \times \spd{n,g}:\left\langle\ominus_{p s d, g} P \oplus_{p s d, g} Q, W \right\rangle^{p s d, g}=0\right\}.
    \end{equation}
    where $\oplus_{p s d, g}$ and $\langle , \rangle^{p s d, g}$ are gyro addition and gyro inner product, which are defined in \cite{nguyen2024matrix}.
\end{definition}

\begin{theorem}(SPSD Pseudo-gyrodistance \cite{nguyen2024matrix})
    \label{app:thm:spsd_gyrodistance}
    Let $W=\left(U_W, S_W\right), P=$ $\left(U_P, S_P\right), X=\left(U_X, S_X\right) \in \grasonb{p,n} \times \spd{n,g}$, and $\mathcal{H}_{A, P}^{p s d, g}$ be a hypergyroplane in structure space $\grasonb{p,n} \times \spd{n,g}$. Then the pseudo-gyrodistance from $X$ to $\mathcal{H}_{A, P}^{p s d, g}$ is given by
    \begin{equation}
        \label{eq:gyro_spsd_mlr}
        \bar{d}\left(X, H_{W, P}^{psd, g}\right)=\frac{\left|\lambda\left\langle\left(\widetilde{\ominus}_{g r} U_P \widetilde{\oplus}_{g r} U_X\right)\left(\widetilde{\ominus}_{g r} U_P \widetilde{\oplus}_{g r} U_X\right)^T, U_W U_W^T\right\rangle^{g r}+\left\langle\ominus_g S_P \oplus_g S_X, S_W\right\rangle^g\right|}{\sqrt{\lambda\left(\left\|U_W U_W^T\right\|^{g r}\right)^2+\left(\left\|S_W\right\|^g\right)^2}},
    \end{equation}
    where $\|.\|^{g r}$ and $\|.\|^g$ are the gyro norms  on the Grassmann and SPD \cite{nguyen2024matrix}, and $\langle , \rangle^{gr}$ and $\langle , \rangle^{g}$ are gyro inner products \cite{nguyen2024matrix}. $\widetilde{\oplus}_{g r}$ and $\oplus_g$ are gyro additions on $\grasonb{p,n}$ and $\spd{p,g}$.
\end{theorem}

Denoting $g^{gr}$ as the canonical metric on $\grasonb{p,n}$ and $g$ as AIM, LEM, or LCM, we can prove that \cref{app:thm:spsd_gyrodistance} is the special case of our \cref{thm:rie_margin_dist}.

\begin{theorem} \label{app:thm:equivalence_rmlr_gyro}
    Under the product metric $g^{psd,g} = \lambda g^{gr} \times g$, 
    the Riemannian hyperplane in \cref{eq:r_hyperplane} on the SPSD manifold equals the SPSD hypergyroplane in \cref{app:def:spsd_hyperplane}.
    Similarly, the Riemannian margin distance in \cref{thm:rie_margin_dist} on the SPSD manifold equals SPSD Pseudo-gyrodistance in \cref{app:thm:spsd_gyrodistance}.
\end{theorem}

\begin{proof}
    Following the notations in \cref{app:def:spsd_hyperplane,app:thm:spsd_gyrodistance}, we further denote $P = U_P S_P U_P^\top$, $Q = U_Q S_Q U_Q^\top$, $W = U_W S_W U_W^\top$, and $X = U_X S_X U_X^\top$ with $U_P, U_Q, U_W, U_X \in \grasonb{p,n}$ and $S_P, S_Q, S_W, S_X \in \spd{p,g}$.
    $I_p$ is the $p \times p$ identity matrix.
    $\idonb=(I_{p},0)^\top$ is the gyro identity on $\grasonb{p,n}$. $\widetilde{\Gamma}^{gr}$, $\widetilde{\rielog}^{gr}$, and $\langle , \rangle^{gr}_{U_p}$ are Riemannian parallel transport along a geodesic, logarithm and Riemannian metric at $U_p$ on $\grasonb{p,n}$.
    $\Gamma^{g}$, $\rielog^{g}$, and $\langle , \rangle^{g}_{S_p}$ are Riemannian parallel transport along a geodesic, logarithm and Riemannian metric at $S_p$ on $\spd{p,g}$.
    $\Gamma^{psd,g}$, $\rielog^{psd, g}$, and $\langle , \rangle^{psd, g}_{X}$ are Riemannian parallel transport along a geodesic, logarithm and Riemannian metric at $X$ on $\spsd{n,p} \cong \grasonb{p,n} \times \spd{p}$.
    
    First, we show that the SPSD hypergyroplane equals our Riemannian hyperplane in \cref{eq:r_hyperplane}.
    We have the following
    \begin{equation}
    \label{app:eq:gyro_hyerp_as_riem_hyper}
    \begin{aligned}
        &\left\langle \spsdominus P \spsdoplus Q, W \right\rangle^{p s d, g} \\
        &\stackrel{(1)}{=} \lambda \langle \grassominus U_P \grassoplus U_Q, U_W  \rangle^{gr} + \langle \gominus S_P \goplus S_Q, S_W \rangle^{g} \\
        &\stackrel{(2)}{=} \lambda \langle \rielog^{gr}_{U_P} U_Q, A_{U_W} \rangle^{gr}_{U_P}
        + \langle \rielog^{gr}_{S_P} S_Q, A_{S_W} \rangle^{g}_{S_P} \\
        &\stackrel{(3)}{=} \langle \rielog^{psd, g}_P Q, \tilde{A} \rangle^{psd,g}_P
    \end{aligned}
    \end{equation}
    where $\tilde{A}_{U_W}= \widetilde{\Gamma}^{gr}_{\idonb \rightarrow U_P} \left( \widetilde{\rielog}^{gr}_{\idonb} (U_W) \right)$, $\tilde{A}_{S_W}= \Gamma^{g}_{I_p \rightarrow S_P} \left( \rielog^{g}_{I_p} (S_W) \right)$ and $\tilde{A}= (\tilde{A}_{U_W}, \tilde{A}_{S_W}) \in T_P\spsd{n,p} \cong T_{U_P}\grasonb{p,P} \otimes T_{S_P}\spd{p,g}$ with $\otimes$ as the Cartesian product.
    The above derivation comes from the following.
    \begin{enumerate}[label=(\arabic*)]
        \item 
        The definition of gyro addition, gyro inverse, and gyro inner product on the SPSD manifold \cite[Sec. 3.3]{nguyen2024matrix}.
        \item 
        The proof of \cite[Prop. 3.2]{nguyen2024matrix} indicates that similar results also hold on the Grassmannian. Combining Prop. 3.2 and its counterparts in the Grassmannian, one can obtain the equation.
        \item 
        The Riemannian product geometry.
    \end{enumerate}
    By the product geometry of the SPSD manifold, we can immediately get
    \begin{equation} \label{app:eq:tilde_A}
        \tilde{A}
        = (\tilde{A}_{U_W}, \tilde{A}_{S_W})
        =\Gamma^{psd,g}_{\idpp \rightarrow P } \left( \rielog_{\idpp}^{psd,g} \left( W \right) \right) 
    \end{equation}
    where $\idpp=\idonb \idonb^\top$ is the gyro identity on the SPSD manifold.
    
    Next, we show the equivalence between SPSD pseudo-gyrodistance and our Riemannian margin distance:
    \begin{equation}
        \begin{aligned}
            \bar{d}\left(X, H_{W, P}^{psd, g}\right)
            &\stackrel{(1)}{=} \frac{\left|\left\langle \spsdominus P \spsdoplus X, W \right\rangle^{p s d, g} \right|}{ \left \| W \right\|^{psd,g}} \\
            &\stackrel{(2)}{=} \frac{\left| \langle \rielog^{psd, g}_P X, \tilde{A} \rangle^{psd,g}_P \right|}{ \left \| W \right\|^{psd,g}} \\
            &\stackrel{(3)}{=} \frac{\left| \langle \rielog^{psd, g}_P X, \tilde{A} \rangle^{psd,g}_P \right|}{ \left \| \rielog_{\idpp}^{psd,g} \left( W \right) \right\|_{\idpp}^{psd,g}}\\
            &\stackrel{(4)}{=} \frac{\left| \langle \rielog^{psd, g}_P X, \tilde{A} \rangle^{psd,g}_P \right|}{\left \| \Gamma^{psd,g}_{\idpp \rightarrow P } \left(\rielog_{\idpp}^{psd,g} \left( W \right) \right) \right\|_{P}^{psd,g}} \\ 
            &\stackrel{(5)}{=} \frac{\left| \langle \rielog^{psd, g}_P X, \tilde{A} \rangle^{psd,g}_P \right|}{ \left \| \tilde{A} \right\|_{P}^{psd,g}} \\   
            &\stackrel{(6)}{=} d(S,\tilde{H}_{\tilde{A}, P}) \\ 
        \end{aligned}
    \end{equation}
    \begin{enumerate}[label=(\arabic*)]
        \item 
        The definition of gyro addition, gyro inverse, gyro inner product, and gyro norm on the SPSD manifold.
        \item 
        \cref{app:eq:gyro_hyerp_as_riem_hyper}.
        \item 
        The definition of SPSD gyro norm \cite{nguyen2024matrix}.
        \item 
        Riemannian parallel transportation maintains the norm of the tangent vector \cite[Def. 3.1]{do1992riemannian}
        \item 
        \cref{app:eq:tilde_A}
        \item 
        \cref{thm:rie_margin_dist}
    \end{enumerate}
\end{proof}

\begin{remark}  
    We make the following remark w.r.t. gyro and our MLR on the SPSD manifold.
    \begin{enumerate}
        \item 
        \cref{app:eq:tilde_A} indicates that when generating $\tilde{A}$ in our RMLR by parallel transporting a tangent vector $A \in T_{\idpp}\spsd{n,p}$, $\tilde{A}$ is the initial velocity of $W$ in \cref{eq:gyro_spsd_mlr}.
        \item
        Putting pseudo-gyrodistance and Riemannian margin distance into \cref{eq:rmlr_v1}, one can get gyro MLR and our Riemannian MLR. 
        Therefore, \cref{app:thm:equivalence_rmlr_gyro} indicates the equivalence of the gyro MLR with our RMLR on the SPSD.
        \item
        As $g$ are required to induce gyro structures, the metric $g$ in gyro SPSD MLR is confined within AIM, LEM, and LCM.
        However, our SPSD MLR can be the product space of the Grassmannian and SPD manifold under other metrics, such as BWM and PEM, as our framework does not require gyro structures.
    \end{enumerate}
\end{remark}

\section{Theories on the deformed metrics}

\subsection{Limiting cases of the deformed metrics} 
\label{app:lim_cases_deformed_metrics}

\citet{thanwerdas2019affine} generalized $\biparamAIM$ into three-parameters families of metrics by power deformation, \ie $\triparamAIM$.
The family of $\triparamAIM$ comprises $\biparamAIM$ for $\theta = 1$ and approaches $\biparamLEM$ with $\theta \rightarrow 0$ \cite{thanwerdas2019affine}.

\citet{chen2024liebn} extended LCM and $\biparamLEM$ into power-deformed metrics, denoted as $\triparamLEM$ and $\paramLCM$.
The authors show that $\triparamLEM$ is equal to $\biparamLEM$, and $\paramLCM$ interpolates between $\tilde{g}$-LEM ($\theta \rightarrow 0$) and LCM ($\theta=1$), with $\tilde{g}$-LEM defined as
\begin{equation}
    \langle V_1,V_2 \rangle_P = \frac{1}{2} \langle \widetilde{V_1}, \widetilde{V_2} \rangle -\frac{1}{4} \langle \bbD(\widetilde{V_1}), \bbD(\widetilde{V_2}) \rangle, \forall V_i \in T_P\spd{n},
\end{equation}
where $\widetilde{V_i} = \log_{*,P}(V_i)$ with $\log_{*,P}$ as the differential map of matrix logarithm, and $\bbD(V_i)$ is a diagonal matrix consisting of the diagonal elements of $V_i$.

\citet{thanwerdas2022geometry} identified the Alpha-Procrustes metric \cite{minh2022alpha} with power-deformed BWM, denote as $2\theta$-BWM. 
Similarly, $2\theta$-BWM becomes BWM with $\theta=0.5$ \cite{thanwerdas2022geometry}. 
We further show the limiting case of $2\theta$-BWM under $\theta \rightarrow 0$.

\begin{proposition} \label{prop:deformed_bwm_limit}
    $2\theta$-BWM tends to be $(\frac{1}{4},0)$-LEM with $\theta \rightarrow 0$.
\end{proposition} 

Before starting the proof, we first recall a well-known property of deformed metrics \citep{thanwerdas2022geometry}.

\begin{lemma} \label{lem:diformed_metrics_lim}
    Let $\frac{1}{\theta^2}\phi_{\theta}^*g$ be the deformed metric on SPD manifolds pulled back from $g$ by the matrix power $\phi_\theta$ and scaled by $\frac{1}{\theta^2}$.
    Then when $\theta$ tends to 0, for all $P \in \spd{n}$ and all $V \in T_P\spd{n}$, we have 
    \begin{equation} \label{eq:deformed_g_lim}
        (\frac{1}{\theta^2}\phi_{\theta}^*g)_P(V,V) \to g_{I}(\log_{*,P}(V),\log_{*,P}(V)).
    \end{equation}
\end{lemma}

Now, we present our proof for the limiting cases of deformed metrics.
\begin{proof}[Proof of \cref{prop:deformed_bwm_limit}]
    First, we have 
    \begin{equation}
        g_I^{\text{BWM}}(V,V) = \frac{1}{4}\langle V, V\rangle.
    \end{equation}

    By \cref{lem:diformed_metrics_lim}, we have the following:
    \begin{equation}
        \begin{aligned}
            \gparamBWM_P(V,V) 
            &\xrightarrow{\theta \rightarrow 0} g^{\text{BWM}}_I \left(\log_{*,P}(V),\log_{*,P}(V) \right)\\
            &= \frac{1}{4} \langle \log_{*,P}(V), \log_{*,P}(V) \rangle \\
            &= g_P^{(\frac{1}{4},0)\text{-LEM}}\left(V,V \right).
        \end{aligned}
    \end{equation}
\end{proof}

\subsection{Proof of the properties of the deformed metrics \texorpdfstring{(\cref{tab:properties_rie_metrics})}{}}
\label{app:props_deformed_metrics}
In this subsection, we prove the properties presented in \cref{tab:properties_rie_metrics}.
We first present a useful lemma and then present our detailed proof.
This lemma will be useful in the proof of our SPD MLRs as well.

\begin{lemma} \label{lem:scale_metric_rie_opt}
    Supposing a Riemannian manifold $\{ \calM, g \}$ and a positive real scalar $a>0$, the scaling metric $ag$ over $\calM$ shares the same Riemannian logarithmic \& exponential maps and parallel transportation with $g$.
\end{lemma}
\begin{proof}
    Since the Christoffel symbols of $ag$ are identical to those of $g$, the geodesics and parallel transportation under both $ag$ and $g$ remain unchanged.
    The equivalence of geodesics implies that the Riemannian exponential maps are the same for $ag$ and $g$.
    As the inverse of the Riemannian exponential maps, the Riemannian logarithm maps under $ag$ and $g$ are also identical.
\end{proof}

According to \cref{lem:scale_metric_rie_opt}, the geodesic completeness is independent of the scaling factor $a>0$.
By the definition of $\orth{n}$-, left-, right-, and bi-invariance, these invariant properties are also independent of the scaling factor $a>0$.
Without loss of generality, we will omit the scaling factor in the following proof.

\begin{proof}
    Firstly, we prove $\orth{n}$-invariance of $(\theta,\alpha,\beta)$-LEM, $(\theta,\alpha,\beta)$-EM, $(\theta,\alpha,\beta)$-AIM, and $2\theta$-BWM.
    Since the differential of $\phi_{\theta}$ is $\orth{n}$-equivariant, and $\alphabeta$-LEM, $\alphabeta$-EM, $\alphabeta$-AIM, and BWM are $\orth{n}$-invariant \citep{thanwerdas2023n},  $\orth{n}$-invariance are thus acquired.

    Next, we focus on geodesic completeness.
    It can be easily proven that Riemannian isometries preserve geodesic completeness.
    On the other hand, $\alphabeta$-LEM, $\alphabeta$-AIM, and LCM are geodesically complete \citep{thanwerdas2023n,lin2019riemannian}.
    As a direct corollary, geodesic completeness can be obtained since $\phi_{\theta}$ is a Riemannian isometry.

    Finally, we deal with Lie group invariance.
    Similarly, it can be readily proved that Lie group invariance is preserved under isometries.
    LCM, LEM, and $\alphabeta$-AIM are Lie group bi-invariant \citep{lin2019riemannian}, bi-invariant \citep{arsigny2005fast}, and left-invariant \citep{thanwerdas2022theoretically}.
    As an isometric pullback metric from the standard LEM \citep{thanwerdas2023n}, $\alphabeta$-LEM is, therefore, Lie group bi-invariant.
    As pullback metrics, $(\theta,\alpha,\beta)$-LEM, $(\theta,\alpha,\beta)$-AIM, and $\theta$-LCM are therefore bi-invariant, left-invariant, and bi-invariant, respectively.
\end{proof}

\section{Computational details on the SPD MLR under power-deformed BWM}
\label{app:sec:detail_bwm_spdmlr}

\subsection{Matrix square roots in the SPD MLR under power-deformed BWM}
In the case of MLRs induced by $2\theta$-BWM, computing square roots like $(BA)^{\frac{1}{2}}$ and $(AB)^{\frac{1}{2}}$ with $B,A \in \spd{n}$ poses a challenge.
    Eigendecomposition cannot be directly applied since $BA$ and $AB$ are no longer symmetric, let alone positive definitity. 
    Instead, we use the following formulas to compute these square roots \citep{minh2022alpha}:
    \begin{equation}
        (BA)^{\frac{1}{2}}=B^{\frac{1}{2}}(B^{\frac{1}{2}}A B^{\frac{1}{2}})^{\frac{1}{2}}B^{-\frac{1}{2}} \text{ and } (AB)^{\frac{1}{2}}=[(BA)^{\frac{1}{2}}]^\top,
    \end{equation}
    where the involved square roots can be computed using eigendecomposition or singular value decomposition (SVD).

\subsection{Numerical stability of the SPD MLR under power-deformed BWM} 
Let us first explain why we abandon parallel transportation on the SPD MLR derived from $2\theta$-BWM. 
Then, we propose our numerically stable methods for computing the SPD MLR based on $2\theta$-BWM.

\subsubsection{Instability of parallel transportation under power-deformed BWM}
\label{app:num_stab_bwm_mlr}
As discussed in \cref{thm:rmlr}, there are two ways to generate $\tilde{A}$ in SPD MLR: parallel transportation and Lie group translation.
However, parallel transportation under $2\theta$-BWM could cause numerical problems.
W.l.o.g., we focus on the standard BWM as $2\theta$-BWM is isometric to the BWM.

Although the general solution of parallel transportation under BWM is the solution of an ODE, for the case of parallel transportation starting from the identity matrix, we have a closed-form expression \citep{thanwerdas2023n}:
\begin{equation} \label{eq:pt_bwm_from_id}
    \Gamma_{I \rightarrow P}(V)= U\left[\sqrt{\frac{\sigma_i+\sigma_j}{2}}\left[U^{\top} V U\right]_{i j}\right] U^{\top},
\end{equation}
where $P=U \Sigma U^\top$ is the eigendecomposition of $P \in \spd{n}$.
There would be no problem in the forward computation of \cref{eq:pt_bwm_from_id}.
However, during backpropagation (BP), \cref{eq:pt_bwm_from_id} would require the BP of eigendecomposition, involving the calculation of $\nicefrac{1}{(\sigma_i-\sigma_j)}$ \citep[Prop. 2]{ionescu2015matrix}.
When $\sigma_i$ is close to $\sigma_j$, the BP of eigendecomposition could be problematic.

\subsubsection{Numerically stable methods for the SPD MLR under power-deformed BWM}
\label{app:bp_Ly}
To bypass the instability of parallel transportation under BWM, we use Lie group left translation to generate $\tilde{A}$ in MLRs induced from $2\theta$-BWM.
However, there is another problem that could cause instability.
The computation of the Riemannian metric of $2\theta$-BWM requires solving the Lyapunov operator, \ie $\calL_P[V] P+ P \calL_P[V]=V$.
Under the case of symmetric matrices, the Lyapunov operator can be obtained by eigendecomposition:
\begin{equation} \label{eq:solution_ly_sym}
    \calL_P[V] = U\left[\frac{V_{i j}^{\prime}}{\sigma_i+\sigma_j}\right]_{i, j} U^{\top},
\end{equation}
where $V \in \sym{n}$, $UV'U^\top=V$, and $P=U\Sigma U^\top$ is the eigendecomposition of $P \in \spd{n}$.
Similar with \cref{eq:pt_bwm_from_id}, the BP of \cref{eq:solution_ly_sym} requires $\nicefrac{1}{(\sigma_i-\sigma_j)}$, undermining the numerical stability.

To remedy this problem, we proposed the following formula to stably compute the BP of \cref{eq:solution_ly_sym}.

\begin{proposition}
    For all $P \in \spd{n}$ and all $V \in \sym{n}$, we denote the Lyapunov equation as
    \begin{equation} \label{eq:ly_symbol}
        XP+PX=V,
    \end{equation}
    where $X=\calL_P[V]$.
    Given the gradient $\frac{\partial L}{\partial X}$ of loss $L$ w.r.t. $X$, then the BP of the Lyapunov operator can be computed by:
    \begin{align}
        \frac{\partial L}{\partial V} &= \calL_P[\frac{\partial L}{\partial X}],\\
        \frac{\partial L}{\partial P} &= -X\calL_P[\frac{\partial L}{\partial X}]-\calL_P[\frac{\partial L}{\partial X}]X,
    \end{align}
\end{proposition}
where $\calL_P[\cdot]$ can be computed by \cref{eq:solution_ly_sym}.
\begin{proof}
    Differentiating both sides of \cref{eq:ly_symbol}, we obtain
    \begin{align}
        & \diff X P + X \diff P + \diff P X+P \diff X=\diff V, \\
        \implies & \diff X P +P \diff X = \diff V - X \diff P - \diff P X,\\
        \implies & \diff X = \calL_P[\diff V - X \diff P - \diff P X].
    \end{align}
    Besides, easy computations show that
    \begin{equation}
        \calL_P[V] : W = V : \calL_P[W], \forall W,V \in \sym{n},
    \end{equation}
    where $\cdot : \cdot$ denotes the standard Frobenius inner product.

    Then we have the following:
    \begin{align}
        & \frac{\partial L}{\partial X} : \diff X = \frac{\partial L}{\partial X} : \calL_P[\diff V - X \diff P - \diff P X],\\
        \implies & \frac{\partial L}{\partial X} : \diff X = \calL_P[\frac{\partial L}{\partial X}] : \diff V + \left( -X\calL_P[\frac{\partial L}{\partial X}]-\calL_P[\frac{\partial L}{\partial X}]X \right): \diff P.
    \end{align}
\end{proof}
\begin{remark}
    \cref{eq:solution_ly_sym} needs to be computed in the Lyapunov operator's forward and backward process.
    Therefore, in the forward process, we can save the intermediate matrices $U$ and $K$ with $K_{i,j}=\left[\frac{1}{\sigma_i+\sigma_j}\right]_{i, j}$, and then use them to compute the backward process efficiently.
\end{remark}

\section{Implementation details and additional experiments}
\label{app:sec:exp_details}
This section offers additional details on the experiments of SPD and Lie MLRs.

\subsection{Additional details and experiments on the SPD MLRs}

\subsubsection{Basic layers in SPDNet and TSMNet} \label{app:spdnet_spddsmbn}

SPDNet \citep{huang2017riemannian} is the most classic SPD neural network. 
SPDNet mimics the conventional densely connected feedforward network, consisting of three basic building blocks:
\begin{align}
    &\text{BiMap layer: }  S^{k} = W^k S^{k-1} W^{k \top}, \text { with } W^k \text { semi-orthogonal,}\\
    &\text{ReEig layer: } S^{k}=U^{k-1} \max (\Sigma^{k-1}, \epsilon I_{n}) U^{k-1 \top},
    \text { with } S^{k-1}=U^{k-1} \Sigma^{k-1} U^{k-1 \top},\\
    &\text{LogEig layer: } S^{k}=\log(S^{k-1}).
\end{align}
where $\max()$ is element-wise maximization.
BiMap and ReEig mimic transformation and non-linear activation, while LogEig maps SPD matrices into the tangent space at the identity matrix for classification.

SPDNetBN \citep{brooks2019riemannian} further proposed Riemannian batch normalization based on AIM:
\begin{align}
    & \text{ Centering from geometric mean } \mathfrak{G} : \forall i \leq N, \bar{S}_i=\mathfrak{G}^{-\frac{1}{2}} S_i \mathfrak{G}^{-\frac{1}{2}},\\
    & \text{ Biasing towards SPD parameter } G: \forall i \leq N, \tilde{S}_i=G^{\frac{1}{2}} \bar{S}_i G^{\frac{1}{2}}.
\end{align}

SPD domain-specific momentum batch normalization (SPDDSMBN) \cite{kobler2022spd} is an improved version of SPDNetBN.
Apart from controlling the mean, it can also control variance.
The key operation in SPDDSMBN of controlling mean and variance is:
\begin{equation}
    \Gamma_{I \rightarrow G} \circ \Gamma_{\mathfrak{G} \rightarrow I}(S_i)^{\frac{\nu}{\bar{\nu}+\varepsilon}},
\end{equation}
where $\mathfrak{G}$ and $\bar{v}$ are Riemannian mean and variance.
Inspired by \cite{yong2020momentum}, during the training stage, SPDDSMBN generates running means and running variances for training and testing with distinct momentum parameters.
Besides, it sets $\mathfrak{G}$ and $\bar{v}$ as the running mean and running variance w.r.t. training for training and the ones w.r.t. testing for testing.
SPDDSMBN also applies domain-specific techniques \citep{chang2019domain}, keeping multiple parallel BN
layers and distributing observations according to the associated domains.
To crack cross-domain knowledge, $v$ is uniformly learned across all domains, and $G$ is set to be the identity matrix.
TSMNet \cite{kobler2022spd} adopted SPDDSMBN to solve domain adaptation in EEG classification.

In the above models, the Euclidean MLR in the co-domain of matrix logarithm (matrix logarithm + FC + softmax) is used for classification.
Following the terminology in \cite{chen2024spdrmlr}, we call this classifier as \textbf{LogEig MLR}.
The LogEig MLR is the Euclidean classifier in the tangent space at the identity, which might distort the innate geometry of the SPD manifold.

\subsubsection{Datasets and preprocessing}
\label{app:subsubsec:datasets_spd}

\textbf{Radar\footnote{https://www.dropbox.com/s/dfnlx2bnyh3kjwy/data.zip?dl=0}: }
This dataset \citep{brooks2019riemannian} consists of 3,000 synthetic radar signals.
Following the protocol in \cite{brooks2019riemannian}, each signal is split into windows of length 20, resulting in 3,000 SPD covariance matrices of $20 \times 20$ equally distributed in 3 classes.

\textbf{HDM05\footnote{https://resources.mpi-inf.mpg.de/HDM05/}: }
This dataset \citep{muller2007documentation} contains 2,273 skeleton-based motion capture sequences executed by various actors.
Each frame consists of 3D coordinates of 31 joints of the subjects, and each sequence can be, therefore, modeled by a $93 \times 93$ covariance matrix.
Following the protocol in \cite{brooks2019riemannian}, we trim the dataset down to 2086 sequences scattered throughout 117 classes by removing some under-represented classes.

\textbf{Hinss2021\footnote{https://zenodo.org/record/5055046}: }
This dataset \citep{hinss_eegdata_2021} is a recent competition dataset consisting of EEG signals for mental workload estimation.
The dataset is used for two types of experiments: inter-session and inter-subject, which are modeled as domain adaptation problems.
% Although EEG can measure brain activity from the human scalp, low signal-to-noise ratio (SNR), domain shifts, and low specificity in EEG signals render statistical learning a challenging task \citep{wolpaw2002brain}.
Recently, geometry-aware methods have shown promising performance in EEG classification \citep{yair2019parallel,kobler2022spd}.
We choose the SOTA method, TSMNet \citep{kobler2022spd}, as our baseline model on this dataset.
We follow the Python implementation\footnote{https://github.com/rkobler/TSMNet} \citep{kobler2022spd} to carry out preprocessing.
In detail, the python package MOABB \citep{jayaram_moabb_2018} and MNE \citep{gramfort_meg_2013} are used to preprocess the datasets.
The applied steps include resampling the EEG signals to 250/256 Hz, applying temporal filters to extract oscillatory EEG activity in the 4 to 36 Hz range, extracting short segments ( $\leq 3$s) associated with a class label, and finally obtaining $40 \times 40 $ SPD covariance matrices.

\textbf{Disease \cite{anderson1991infectious}: }
It represents a disease propagation tree, simulating the SIR disease transmission model~\cite{anderson1991infectious}, with each node representing either an infection or a non-infection state.

\textbf{Cora \cite{sen2008collective}: }
It is a citation network where nodes represent scientific papers in the area
of machine learning, edges are citations between them, and node labels are academic (sub)areas.

\textbf{Pubmed \cite{namata2012query}: }
This is a standard benchmark describing citation networks where nodes represent scientific papers in the area of medicine, edges are citations between them, and node labels are academic (sub)areas.

For the Disease, Cora and Pubmed datasets, we follow \cite{zhao2023modeling} to model features into $\spd{3}$.

\subsubsection{Implementation details}
\label{app:subsubsec:implementaion_details}

\textbf{SPDNet \citep{huang2017riemannian} and TSMNet \citep{kobler2022spd}: }
We follow the official Pytorch code of
SPDNetBN\footnote{https://proceedings.neurips.cc/paper\_files/paper/2019/file/6e69ebbfad976d4637bb4b39de261bf7-Supplemental.zip}
and
TSMNet\footnote{https://github.com/rkobler/TSMNet} to implement our experiments.
To evaluate the performance of our intrinsic classifiers, we substitute the LogEig MLR in SPDNet and TSMNet with our SPD MLRs.
We implement our SPD MLRs induced from five parameterized metrics.
On the Radar and HDM05 datasets, the learning rate is $1e^{-2}$, and the batch size is 30.
On the Hinss2021 dataset, following \cite{kobler2022spd}, the learning rate is $1e^{-3}$ with a $1e^{-4}$ weight decay, and batch size is 50. 
The maximum training epoch is 200, 200, and 50, respectively.
We use the standard-cross entropy loss as the training objective and optimize the parameters with the Riemannian AMSGrad optimizer \citep{becigneul2018riemannian}.

\textbf{RResNet \cite{katsman2023riemannian}: } 
We focus on the AIM-based RResNet, and use the official code\footnote{https://github.com/CUAI/Riemannian-Residual-Neural-Networks} and suggested network settings to implement the experiments on the RResNet.
We conduct 10-fold and 5-fold experiments on the HDM05 and NTU datasets. 
Since RResNet is developed based on SPDNet, we use the same learning settings with the SPDNet for the action recognition task, and borrow the best $(\theta,\alpha,\beta)$ from \cref{tb:results_hdm05} for our SPD MLRs under the RResNet backbone. 

\textbf{SPDGCN \citep{zhao2023modeling}: }
We use the official code\footnote{https://github.com/andyweizhao/SPD4GNNs} and the suggested network settings in \citep{zhao2023modeling}. Note that the SPDGCN with SPD MLR remains the same network settings as the vanilla SPDGCN. 
\cref{tab:hyper_spdgcn} presents the hyperparameters $(\theta,\alpha,\beta)$ on different datasets.

\begin{table}[htbp]
  \centering
  \caption{$(\theta,\alpha,\beta)$ of SPD MLRs on the SPDGCN backbone.}
  \label{tab:hyper_spdgcn}%
    \begin{tabular}{cccccc}
    \toprule
    Datasets & $\triparamAIM$ & $\triparamEM$ & $\biparamLEM$ & $\paramBWM$ & $\paramLCM$ \\
    \midrule
    Disease & (0.25,1,0) & (0.25,1,0) & (1,1) & 0.25  & 0.5 \\
    Cora  & (0.5,1,0) & (0.25,1,$\nicefrac{1}{9}$) & (1,$\nicefrac{1}{9}$) & 0.25  & 0.5 \\
    Pubmed & (0.5,1,0) & (0.5,1,0) & (1,-$\nicefrac{1}{3}$) & 0.25  & 0.5 \\
    \bottomrule
    \end{tabular}%
  
\end{table}%

\textbf{Network Architectures:}
We denote the network architecture as $[d_0, d_1,\cdots,d_L]$, where the dimension of the parameter in the $i$-th BiMap layer (\cref{app:spdnet_spddsmbn}) is $d_{i} \times d_{i-1}$. 
For SPDNet, we also validate our SPD MLRs under different network architectures on the Radar and HDM05 datasets.
The network architectures on the Radar dataset are [20, 16, 8] for the 2-block configuration and [20, 16, 14, 12, 10, 8] for the 5-block configuration, while on the HDM05 dataset, the network architectures are [93, 30] for 1-block, [93. 70, 30] for 2-block, and [93, 70, 50, 30] for 3-block.
For TSMNet, the 1-block architecture is [40,20].

\textbf{Scoring Metrics:}
In line with the previous work \citep{brooks2019riemannian,kobler2022spd,zhao2023modeling,katsman2023riemannian}, we use balanced accuracy, the average recall across classes, as the scoring metric for the Hinss2021 dataset, and accuracy for other datasets.
On the Hinss2021 dataset, models are fit and evaluated with a randomized leave 5\% of the sessions (inter-session) or subjects (inter-subject) out cross-validation (CV) scheme.
On other datasets, K-fold experiments are carried out with randomized initialization and split, 

\subsubsection{Hyper-parameters} \label{app:hp_selection}

We implement the SPD MLRs induced by not only five standard metrics, \ie LEM, AIM, EM, LCM, and BWM, but also five families of parameterized metrics.
Therefore, in our SPD MLRs, we have a maximum of three hyper-parameters, \ie $\theta,\alpha,\beta$, where $\alphabeta$ are associated with $\orth{n}$-invariance and $\theta$ controls deformation.
For $\alphabeta$ in $(\theta,\alpha,\beta)$-LEM, $(\theta,\alpha,\beta)$-AIM, and $(\theta,\alpha,\beta)$-EM, recalling \cref{eq:oim_sym}, $\alpha$ is a scaling factors, while $\beta$ measures the relative significance of traces.
As scaling is less important \citep{thanwerdas2019affine}, we set $\alpha=1$.
As for the value of $\beta$, we select it from a predefined set: $\{1,\nicefrac{1}{n},\nicefrac{1}{n^2}, 0, -\nicefrac{1}{n} + \epsilon,-\nicefrac{1}{n^2}\}$, where $n$ is the dimension of input SPD matrices in SPD MLRs. 
The purpose of including $\epsilon \in \bbRplus$ is to ensure $\orth{n}$-invariance ($\alphabeta \in \bfst$).
These chosen values for $\beta$ allow for amplifying, neutralizing, or suppressing the trace components, depending on the characteristics of the datasets.
For the deformation factor $\theta$, we roughly select its value around its deformation boundary, \ie [0.25,1.5] for $(\theta,\alpha,\beta)$-AIM,  [0.5,1.5] for $\theta$-LCM, [0.25,1.5] and $(\theta,\alpha,\beta)$-EM, [0.25,0.75] for $2\theta$-BWM.
The details values are listed in \cref{tab:hyper_param_values}.

\begin{table}[htbp]
    \centering
    \caption{Candidate values for hyper-parameters in SPD MLRs}
    \label{tab:hyper_param_values}%
    % \resizebox{0.99\linewidth}{!}{
    \begin{tabular}{ccccc}
    \toprule
    Metric & $(\theta,\alpha,\beta)$-AIM & $(\theta,\alpha,\beta)$-EM & $\theta$-LCM & $2\theta$-BWM \\
    \midrule
    Candidate Values & \{ 0.25,0.5,0.75,1,1.25,1.5 \} & \{0.5,1,1.5 \} & \{0.5,1,1.5 \} & \{0.25,0.5,0.75 \} \\
    \bottomrule
    \end{tabular}%
    % }
\end{table}%

\subsubsection{Model efficiency} \label{app:efficiency}

\begin{table}[htbp]
    \centering
    \caption{Training efficiency (s/epoch).}
    \label{tab:training_efficiency}
    \begin{tabular}{ccccc}
    \toprule
    \multirow{2}[4]{*}{Methods} & \multirow{2}[4]{*}{Radar} & \multirow{2}[4]{*}{HDM05} & \multicolumn{2}{c}{Hinss2021} \\
    \cmidrule{4-5}          &       &       & Inter-session & Inter-subject \\
    \midrule
    Baseline & 1.36  & 1.95  & 0.18  & 8.31 \\
    AIM-MLR & 1.75  & 31.64 & 0.38  & 13.3 \\
    EM-MLR & 1.34  & 3.91  & 0.19  & 8.23 \\
    LEM-MLR & 1.5   & 4.7   & 0.24  & 10.13 \\
    BWM-MLR & 1.75  & 33.14 & 0.38  & 13.84 \\
    LCM-MLR & 1.35  & 3.29  & 0.18  & 8.35 \\
    \bottomrule
    \end{tabular}
\end{table}

We adopt the deepest architectures, namely [20, 16, 14, 12, 10, 8] for the Radar dataset, [93, 70, 50, 30] for the HDM05 dataset, and [40, 20] for the Hinss2021 dataset.
For simplicity, we focus on the SPD MLRs induced by standard metrics, \ie AIM, EM, LEM, BWM, and LCM.
The average training time (in seconds) per epoch is reported in \cref{tab:training_efficiency}.
Generally, when the number of classes is small (\eg 3 in the Radar and Hinss2021 datasets), our SPD MLRs only bring minor additional training time compared to the baseline LogEig MLR.
However, when dealing with a larger number of classes (\eg 117 classes in the HDM05 dataset), there could be some inefficiency caused by our SPD MLRs. 
This is because each class requires an SPD parameter, and each parameter might require matrix decomposition in the forward or optimization processes during training.
Nonetheless, the SPD MLRs induced by EM or LCM generally achieve comparable efficiency with the vanilla LogEig MLR.
This is due to the fast computation of their Riemannian operators, making them efficient choices for tasks with a larger number of classes.
This result highlights the flexibility of our framework and its applicability to various scenarios.

\begin{figure}[htbp]
\centering
\includegraphics[width=0.6\linewidth,trim=0 0mm 0 0cm]{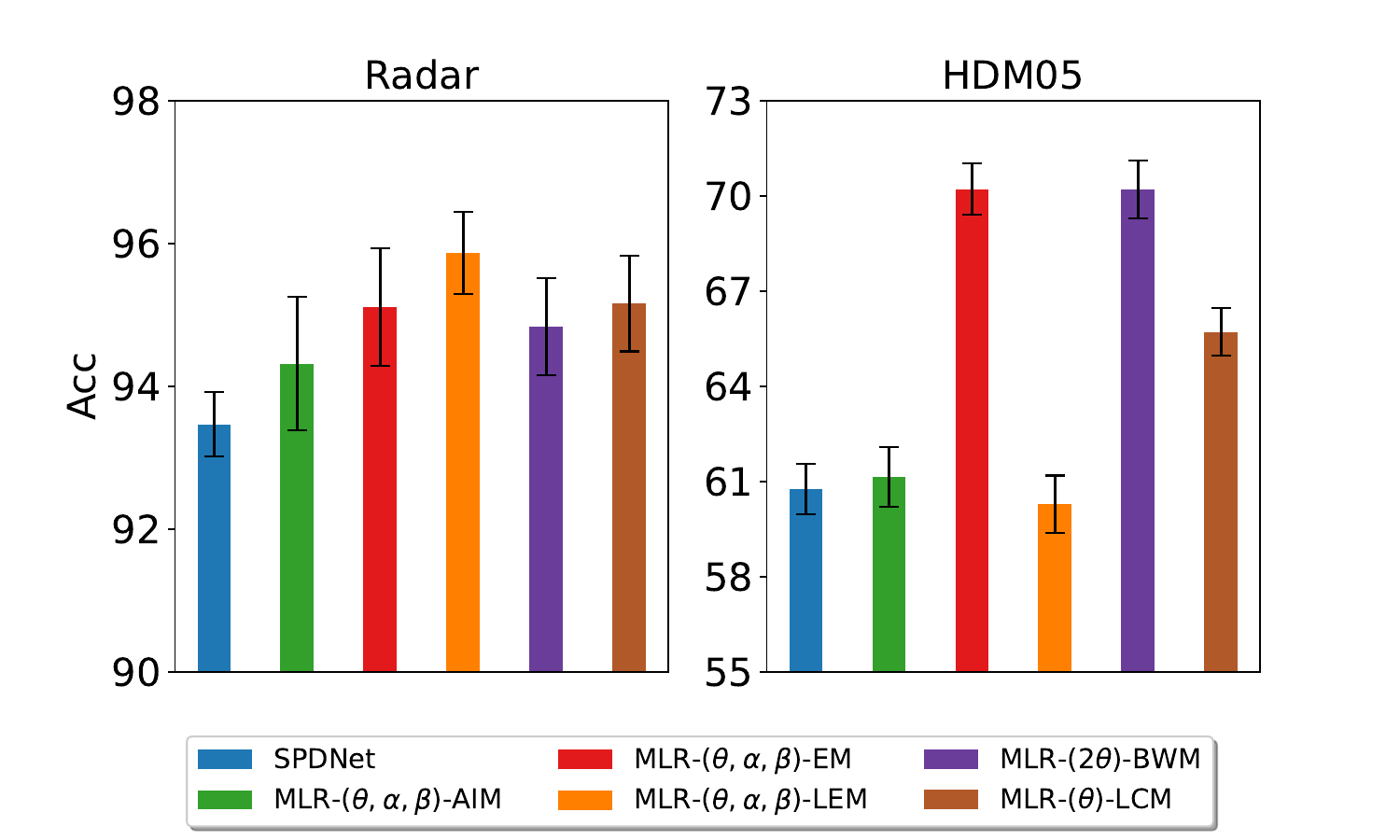}
\caption{Visualization of 10-fold average accuracy of SPDNet with different SPD MLRs on the Radar and HDM05 datasets. The error bar denotes the standard deviation.
}
\label{fig:visulization}
% \vspace{-5mm}
\end{figure}

\subsubsection{Visualization}
\label{app:subsec:visulization}

We visualize the 10-fold average results of SPDNet with different classifiers on the Radar and HDM05 datasets. We focus on the deepest architectures, \ie. [20,16,14,12,10,8] for the Radar dataset, and [93,70,50,30] for the HDM05 dataset. Note that we only report the SPD MLR with the best hyper-parameters $(\theta,\alpha,\beta)$. The figures are presented in \cref{fig:visulization}. All the results are sourced from \cref{tb:results_radar,tb:results_hdm05}.

\subsection{Additional details and experiments on the Lie MLR}

\subsubsection{Basic layers in LieNet}
\label{app:subsec:lienet}
LieNet \cite{huang2017deep} is the most classic neural network on rotation matrices.
The latent space of LieNet is the Lie group $\soprod{N}{3}=\so{3} \times \so{3} \cdots \times \so{3}$, \ie $R=(R_1,\cdots,R_N) \in \soprod{N}{3}$.
The group and manifold structures on $\soprod{N}{3}$ are defined by product spaces. 
For instance, $R^1 \odot R^2 = (R^1_1R^2_1, \cdots, R^1_N R^2_N)$.
There are three basic layers in LieNet:
\begin{align}
    &\text{RotMap layer: }  R^{k} = W^k \odot R^{k-1}, \text { with } W^k \in \soprod{N}{3},\\
    &\text{RotPooling layer: } R^{k}_i=
    \begin{cases}
        R_{m_i, n_i}^{k-1}, & \text { if } \Theta\left(R_{m_i, n_i}^{k-1}\right)>\Theta\left(R_{n_i, m_i}^{k-1}\right), \\ R_{n_i, m_i}^{k-1}, & \text { otherwise, }
    \end{cases}, \\
    &\text{LogMap layer: } R^{k}=\log(R^{k-1}),
\end{align}
where $\Theta(\cdot)$ is the Euler angle, and $(n_i,m_i)$ are two indexes.
The RotMap and RotPooling layers mimic the convolution and pooling layers, while the LogMap layer map rotation matrices into tangent space for classification.
In the official Matlab implementation, the LogMap layer is implemented as the Euler axis-angle representation.
The classification is performed by Euler axis-angle + FC + Softmax.
As the axis-angle is an equivalent representation of matrix logarithm, we call this classifier as \textbf{LogEig MLR} as well.
This classifier is, therefore, also non-intrinsic.

In LieNet, each rotation feature has a shape of [num, frame, 3, 3], where num and frame denote spatial and temporal dimensions. 
The RotPooling layer is performed either along spatial or temporal dimensions, while the RotMap layer is performed along spatial dimensions, \ie $W^k$ with a size of [num, 3, 3].

\subsubsection{Datasets and preprocessing}
\label{app:subsub:datasets_son}

For a fair comparison, we follow LieNet to use G3D \citep{bloom2012g3d} and HDM05 datasets to validate our Lie MLR.

\textbf{G3D\cite{bloom2012g3d}:}
This dataset consists of 663 sequences of 20 different gaming actions.
Each sequence is recorded by 3D locations of 20 joints (i.e., 19 bones).

\textbf{HDM05: }
We trim it down by removing some under-represented sequences, resulting in 2,326 sequences scattered throughout 122 classes.
Following \cite{huang2017riemannian}, we use the code of \cite{vemulapalli2014human} to represent each skeleton sequence as a point on the Lie group $\soprod{N \times T}{3}$, where $N$ and $T$ denote spatial and temporal dimensions.
As preprocessed in \cite{huang2017riemannian}, we set $T$ as 100 and 16 for each sequence on the G3D and HDM05 datasets, respectively.

\subsubsection{Implementation details}

\textbf{LieNet:} 
Note that the official code of LieNet\footnote{https://github.com/zhiwu-huang/LieNet} is developed by Matlab. 
We follow the open-sourced Pytorch code\footnote{https://github.com/hjf1997/LieNet} to implement our experiments.
To reproduce LieNet more faithfully, we made the following modifications to this Pytorch code.
We re-code the LogMap and RotPooling layers to make them consistent with the official Matlab implementation.
In addition, we also extend the existing Riemannian optimization package geoopt \cite{becigneul2018riemannian} into $\so{3}$ to allow for Riemannian version of SGD, ADAM, and AMSGrad on $\so{3}$, which is missing in the current package. 
However, we find that SGD is the best optimizer for LieNet. 
Therefore, we use SGD as our optimizer during the experiments.
% In the original LieNet, the LogMap layer is implemented as the Euler axis-angle representation.
% The classification is performed by Euler axis-angle representation + FC + Softmax.
% This classifier is non-intrinsic.
% We call this classifier the axis-angle classifier.

\textbf{Lie MLR:}
We use our Lie MLR to replace the axis-angle classifier in LieNet and call the resulting network LieNet+LieMLR.
To alleviate the computational burden, we set each $P_k$ as the dimension of [num, 3, 3], where num is the spacial dimension of the input of the Lie MLR layer.
In other words, $P_k$ is shared in the temporal dimension.
We adopt Pytroch3D \cite{ravi2020accelerating} to calculate the matrix logarithm.
Due to the instabilities of \texttt{pytorch3d.transforms.so3\_log\_map}, we use \texttt{pytorch3d.transforms.matrix\_to\_axis\_angle} first to calculate the quaternion axis and angle, and then convert this representation into matrix logarithm\footnote{https://github.com/facebookresearch/pytorch3d/issues/188}.

\textbf{Training Details: }
Following \cite{huang2017deep}, we focus on the 3Blocks and 2Blocks architecture for the G3D and HDM05 datasets, which are the suggested architectures for these two datasets.
The learning rate is $1e^{-2}$ on both datasets, and we further set weight decay as $1e^{-5}$ on the G3D dataset.
For LieNet and LieNet+LieMLR, we use \texttt{torch.nn.utils.clip\_grad\_norm\_} for gradient clipping with a clipping factor of 5.
The clipping is imposed to the dimensionality reduction weight in the final FC linear on LieNet, or, accordingly, $A = \{A_1,\cdots, A_k\}$ in the Lie MLR layer on LieNet+LieMLR.

\textbf{Scoring Metrics:}
For the G3D dataset, following LieNet \cite{huang2017deep}, we adopt a 10-fold cross-subject test setting, where half the subjects are used for training and the other half are employed for testing. 
For the HDM05 dataset, following \cite{huang2017deep}, we randomly select half of the sequences for training and the rest for testing.
Due the instabilities of LieNet, we conduct 20-fold experiments and select the best 10 folds to evaluate the performance.

\subsection{Hardware}
\label{app:subsec:hardware}
All experiments use an Intel Core i9-7960X CPU with 32GB RAM and an NVIDIA GeForce RTX 2080 Ti GPU.

\section{Proofs}
\label{app:proof}

\subsection{Proof of \cref{thm:rie_margin_dist}}
\linkofproof{thm:rie_margin_dist}

\begin{proof}[Proof of \cref{thm:rie_margin_dist}]
    Let us first solve $Y^*$ in \cref{eq:margin_dist_reform_v2}, which is the solution to the following constrained optimization problem:
    \begin{equation} \label{eq:q_problem_v1}
        \underset{Y}{\max} \left(\frac{\langle \rielog_P Y, \rielog_P S \rangle_P}{\|\rielog_P Y\|_P,\|\rielog_P S\|_P} \right)
        \quad
        \st \langle \rielog_{P} S, \tilde{A} \rangle_{P}=0
    \end{equation}
    Note that \cref{eq:q_problem_v1} is well-defined due to the existence of the Riemannian logarithm.
    Although, \cref{eq:q_problem_v1} is normally non-convex, \cref{eq:q_problem_v1} and \cref{eq:margin_dist_reform_v2} can be reduced to a Euclidean problem:
    \begin{align}
        \label{eq:max_cos_tpm}
        \underset{\tilde{Y}}{\max} \frac{\langle \tilde{Y}, \tilde{S}\rangle_P}{\| \tilde{Y} \|_P\| \tilde{S} \|_P} \quad \st \langle \tilde{Y}, \tilde{A} \rangle_P=0,\\
        \label{eq:dist_S_to_hp_v2}
        d(S,\tilde{H}_{\tilde{A}, P})=\sin (\angle SPY^*) \| \tilde{S} \|_P,
    \end{align}
    where $\tilde{Y}=\rielog_P Y$ and $\tilde{S}=\rielog_P S$.

    Let us first discuss \cref{eq:max_cos_tpm}.
    Denote the solution of \cref{eq:max_cos_tpm} as $\tilde{Y}^*$.
    Note that $\tilde{Y}^*$ is not necessarily unique.
    Note that $\rieexp_P$ is only well-defined locally.
    More precisely, $\rieexp_P$ is well-defined in an open ball $\mathrm{B}_{\epsilon}(0)$ centered at $0 \in T_P\calM$.
    Therefore, $\tilde{Y}^*$ might not be in $\mathrm{B}_{\epsilon}(0)$.
    In this case, we can scale $\tilde{Y}^*$ into $\mathrm{B}_{\epsilon}(0)$, and the scaled $\tilde{Y}^*$ is still the maximizer of \cref{eq:max_cos_tpm}.
    Therefore, w.l.o.g., we assume $\tilde{Y}^* \in \mathrm{B}_{\epsilon}(0)$.

    Putting $\tilde{Y}^*$ into \cref{eq:dist_S_to_hp_v2}, \cref{eq:dist_S_to_hp_v2} is reduced to the distance to the hyperplane $\langle \tilde{Y}, \tilde{A} \rangle_P=0$ in the Euclidean space $\{ T_P\calM, \langle \cdot, \cdot \rangle_P \}$, which has a closed-form solution:
    \begin{align}
        d(S,\tilde{H}_{\tilde{A}, P}) 
        &= \frac{|\langle \tilde{S}, \tilde{A} \rangle_P|}{\| \tilde{A} \|_P},\\
        \label{eq:margin_dist_final}
        &= \frac{|\langle  \rielog_P S, \tilde{A} \rangle _P|}{\| \tilde{A} \|_P}.
    \end{align} 
\end{proof}

\subsection{Proof of \cref{thm:rmlr}}
\linkofproof{thm:rmlr}

\begin{proof}[Proof for \cref{thm:rmlr} \linkofproof{thm:rmlr}]
    Putting the margin distance (\cref{eq:rie_margin_dist}) into \cref{eq:rmlr_v1}, we have the following:
    \begin{equation}
        \begin{aligned}
            p(y=k \mid S)
            &\propto \exp \left(\operatorname{sign}(\langle \tilde{A}_k, \rielog_{P_k}(S) \rangle_{P_k})\|\tilde{A}_k\|_{P_k} d (S, \tilde{H}_{\tilde{A}_k, P_k}) \right),\\
            &= \exp \left(\operatorname{sign}(\langle \tilde{A}_k, \rielog_{P_k}(S) \rangle_{P_k})\|\tilde{A}_k\|_{P_k} \frac{|\langle  \rielog_{P_k} (S), \tilde{A_k} \rangle _{P_k}|}{\| \tilde{A_k} \|_{P_k}} \right),\\
            &= \exp \left( \langle \rielog_{P_k} S,  \tilde{A}_k \rangle_{P_k} \right).
        \end{aligned}
    \end{equation}
\end{proof}

\subsection{Proof of \cref{prop:spd_parametrized_metrics}}
\linkofproof{prop:spd_parametrized_metrics}

\begin{proof}[Proof for \cref{prop:spd_parametrized_metrics} \linkofproof{prop:spd_parametrized_metrics}]
    The Riemannian metric $\biparamEM$ at $I$ is
    \begin{equation}
        g_I^{\alphabeta\text{-EM}}(V,V) = \langle V, V\rangle^{\alphabeta}. 
    \end{equation}
    By \cref{lem:diformed_metrics_lim}, we have the following
    \begin{equation}
        \begin{aligned}
            \gtriparamEM_P(V,V) 
            &\xrightarrow{\theta \rightarrow 0} g_I^{\alphabeta\text{-EM}}\left(\log_{*,P}(V),\log_{*,P}(V) \right)\\
            &= \langle \log_{*,P}(V), \log_{*,P}(V) \rangle^{\alphabeta} \\
            &= g_P^{\alphabeta\text{-LEM}}\left(V,V \right).
        \end{aligned}
    \end{equation}
\end{proof}

\subsection{Proof of \cref{thm:spdmlrs}}
\linkofproof{thm:spdmlrs}

As the five families of metrics presented in \cref{thm:spdmlrs} are pullback metrics, we first present a general result regarding Riemannian MLRs under pullback metrics.

\begin{lemma}[Riemannian MLRs under Pullback Metrics] \label{lem:rmlr_pullback}
    Supposing $\{\calN,g\}$ is a Riemannian manifold and $\phi:\calM \rightarrow \calN$ is a diffeomorphism between manifolds,
    the Riemannian MLR by parallel transportation (\cref{eq:rmlr_final} + \cref{eq:A_by_pt}) on $\calM$ under $\tilde{g}=\phi^* g$ can be obtained by $g$:
    {\small
    \begin{align}
        p(y=k \mid S \in \calM)
        &\propto \exp \left[ \langle \tilde{\rielog}_{P_k} S,  \tilde{\Gamma}_{Q \rightarrow P_k} A_k \rangle_{P_k} \right],\\
        \label{eq:rmlr_pm_pt}
        &= \exp \left [\langle \rielog_{\phi(P_k)} \phi(S), \tilde{A}_k  \rangle_{\phi(P_k)} \right ],
    \end{align}}
    where $\tilde{A}_k = \Gamma_{\phi(Q) \rightarrow \phi(P_k)} \phi_{*,Q} (A_k)$ with $A_k \in T_Q\calM$, $\tilde{\rielog},\tilde{\Gamma}$ are Riemannian logarithm and parallel transportation under $\tilde{g}$, and $\rielog, \Gamma$ are the counterparts under $g$. 
    
    Furthermore, if $\calN$ has a Lie group operation $\odot$, $\calM$ could be endowed with a Lie group structure $\tilde{\odot}$ by $f$. 
    The Riemannian MLR by left translation (\cref{eq:rmlr_final} + \cref{eq:A_by_lt}) on $\calM$ under $\tilde{g}$ and $\tilde{\odot}$ can be calculated by $g$ and $\odot$:
    {\small
    \begin{align}
        \label{eq:rmlr_pm_lt_v0}
        p(y=k \mid S \in \calM)
        &\propto \exp \left[ \langle \tilde{\rielog}_{P_k} S,  \tilde{L}_{\tilde{R}_k *,Q} A_k \rangle_{P_k} \right],\\
        \label{eq:rmlr_pm_lt}
        &= \exp \left[ \langle \rielog_{\phi(P_k)} \phi(S), \tilde{A}_k \rangle_{\phi(P_k)} \right],
    \end{align}}
    where $\tilde{A}_k = L_{R_k *, \phi(Q)} \circ \phi_{*,Q}(A_k)$, $\tilde{R}_k=P_k \tilde{\odot} Q_{\tilde{\odot}}^{-1}$, $R_k=\phi(P) \odot \phi(Q)^{-1}_{\odot}$, and $\tilde{L}_{P_k \tilde{\odot} Q_{\tilde{\odot}}^{-1}}$ is the left translation under $\tilde{\odot}$.
\end{lemma}
\begin{proof}[Proof for \cref{lem:rmlr_pullback}]
    Before starting, we should point out that since $\phi$ is a diffeomorphism, $\tilde{\odot}$ and $\tilde{g}$ are indeed well defined, and $\{\calM, \tilde{g}\}$ forms a Riemannian manifold and $\{\calM, \tilde{\odot} \}$ forms a Lie group.
    We denote $\phi^{-1}_*$ as the differential of $\phi^{-1}$.
    We first focus on the Riemannian MLR by parallel transportation:
    \begin{equation}
        \begin{aligned}
            & p(y=k \mid S \in \calM) \\ 
            &\propto \exp ( \tilde{g}_{P_k} (\tilde{\rielog}_{P_k} S,  \tilde{\Gamma}_{Q \rightarrow P_k} A_k )) \\
            &= \exp \left [g_{\phi(P_k)} \left( \phi_{*,P_k} \circ \phi^{-1}_{*,\phi(P_k)} \rielog_{\phi(P_k)} \phi(S), \phi_{*,P_k} \circ \phi^{-1}_{*,\phi(P_k)} \Gamma_{\phi(Q) \rightarrow \phi(P_k)} \phi_{*,Q} (A_k) \right) \right ] \\
            &= \exp \left [g_{\phi(P_k)}(\rielog_{\phi(P_k)} \phi(S), \Gamma_{\phi(Q) \rightarrow \phi(P_k)} \phi_{*,Q} (A_k)) \right ].
        \end{aligned}
    \end{equation}
    
    In the case of the Riemannian MLR by left translation, we first note that:
    \begin{equation}
        \tilde{L}_{\tilde{R}_k}= \phi^{-1} \circ L_{\phi(P_k) \odot \phi(Q)^{-1}_{\odot}} \circ \phi.
    \end{equation}
    Therefore, the associated differential is:
    \begin{equation} \label{eq:diff_l_pm}
        \tilde{L}_{\tilde{R}_k *}= \phi^{-1}_* \circ L_{\phi(P_k) \odot \phi(Q)^{-1}_{\odot} *} \circ \phi_{*}.
    \end{equation}
    Putting \cref{eq:diff_l_pm} in \cref{eq:rmlr_pm_lt_v0}, we can obtain the results.
\end{proof}

Now, we apply \cref{lem:rmlr_pullback} to derive the expressions of our SPD MLRs presented in \cref{thm:spdmlrs}.
For our cases of SPD MLRs, we set $Q=I$. 
For simplicity, we will omit the subscript $k$ for $P_k$ and $A_k$.
We will first derive the expressions of SPD MLRs under $(\theta,\alpha,\beta)$-LEM, $\theta$-LCM, $(\theta,\alpha,\beta)$-EM, and $(\theta,\alpha,\beta)$-AIM, as they are sourced from \cref{eq:rmlr_pm_pt}.
Then we will proceed to present the expression of MLR under $2\theta$-BWM, which is sourced from \cref{eq:rmlr_pm_lt}.
According to \cref{lem:scale_metric_rie_opt}, the scaled metric $ag$ shares the same Riemannian operators as $g$.
We will use this fact throughout the following proof.

\begin{proof}[Proof of \cref{thm:spdmlrs}]
    For simplicity, we abbreviate $\phi_\theta$ as $\phi$ during the proof. Note that for $2\theta$-BWM, $\phi$ should be understood as $\phi_{2\theta}$.
    We first show $\phi(I)$ and the differential map $\phi_{*, I}$, which will be frequently required in the following proof:
    \begin{align} 
        \phi(I) &= I, \\
        \label{eq:diff_power_diform_map}
        \phi_{ *, I}(A) &= \theta A, \forall A \in T_I\spd{n}.
    \end{align}
    
    Denoting $\phi: \{\spd{n},\tilde{g}\} \rightarrow \{\spd{n}, g\}$, then the SPD MLR under $\tilde{g}$ by parallel transportation with $Q=I$ is
    \begin{equation} \label{eq:rmlr_pm_pt_simplified}
        p(y=k \mid S \in \calM) 
        = \exp \left [g_{\phi(P)}(\rielog_{\phi(P)} \phi(S), \Gamma_{I \rightarrow \phi(P)} \theta A) \right ],
    \end{equation}
    
    Next, we begin to prove the five SPD MLRs one by one.

    \textbf{$(\alpha,\beta)$-LEM:}    
    As shown by \citet{chen2024adaptive}, the standard LEM is the pullback metric from the Euclidean space $\sym{n}$. Similarly, $(\alpha,\beta)$-LEM is also a pullback metric:
    \begin{equation}
        \{\spd{n}, g^{\biparamLEM}\} \stackrel{\log}{\longrightarrow}
        \{\sym{n}, g^{\alphabeta}\}
    \end{equation}
    By \cref{eq:rmlr_pm_pt}, we have
    \begin{align}
        p(y=k \mid S \in \calM) 
        &= \exp \left [ \langle \log(S)-\log(P), \log_{*,I} (A) \rangle^{\alphabeta} \right ]\\
        &= \exp \left [ \langle \log(S)-\log(P), A \rangle^{\alphabeta} \right ].
    \end{align}
    
    \textbf{$\theta$-LCM}:
    Simple computations show that $\theta$-LCM is the scaled pullback metric of standard Euclidean metric in the Euclidean space of lower triangular matrices $\tril{n}$:
    \begin{equation}
        \{ \spd{n}, \theta^2 g^{\paramLCM} \} \stackrel{\phi}{\longrightarrow}
        \{ \spd{n}, g^{\LCM} \} \stackrel{\chol}{\longrightarrow}
        \{ \cho{n}, g^{\text{CM}} \} \stackrel{\dlog}{\longrightarrow}
        \{\sym{n}, g^{\text{E}}\},
    \end{equation}
    where $g^{\text{E}}$ is the standard Frobenius inner product, and $g^{\text{CM}}$ is the Cholesky metric on the Cholesky space $\cho{n}$ \citep{lin2019riemannian}.
    Denoting $\zeta= \dlog \circ \chol \circ \phi$, then we have
    \begin{equation}
        \zeta_{*,I}(A) = \theta \left( \lfloor A \rfloor + \frac{1}{2}\bbD(A) \right), \forall A \in T_I\spd{n}.
    \end{equation}
    Similar with the case of $\triparamLEM$, we have
    \begin{align}
        p(y=k \mid S \in \calM) & \propto \exp \left [ \frac{1}{\theta^2}\langle \zeta(S)-\zeta(P), \zeta_{*,I} A \rangle \right ],\\
        & = \exp \left[ \frac{1}{\theta} \langle \lfloor \tilde{K}\rfloor - \lfloor \tilde{L} \rfloor + \left[\dlog(\bbD(\tilde{K}))-\dlog(\bbD(\tilde{L}))\right], \lfloor A \rfloor + \frac{1}{2}\bbD(A) \rangle \right],
    \end{align}
    where $\tilde{K}=\chol(S^\theta)$, $\tilde{L}=\chol(P^\theta)$, $\bbD(\tilde{K})$ is a diagonal matrix with diagonal elements from $\tilde{K}$, and $\lfloor \tilde{K} \rfloor$ is a strictly lower triangular matrix from $\tilde{K}$.
    
    \textbf{$(\theta,\alpha,\beta)$-EM:}
    Let $\eta=\frac{1}{|\theta|}\phi$.
    Simple computation shows that $\triparamEM$ is the pullback metric of $\biparamEM$:
    \begin{equation}
        \{\spd{n}, g^{\triparamEM}\} 
        \stackrel{\eta}{\longrightarrow}
        \{\spd{n}, g^{\biparamEM}\}.
    \end{equation}
    Besides, we have the following for $\eta$:
    \begin{equation}
        \eta_{*,I}(A) =\sgn \theta A, \forall A \in T_I\spd{n}.
    \end{equation}
    
    According to \cref{eq:rmlr_pm_pt}, we have
    \begin{align}
        p(y=k \mid S \in \calM) 
        &\propto \exp \left[ 
        \langle \eta(S)-\eta(P),\sgn(\theta) A \rangle \right],\\
        &= \exp \left[ \frac{1}{\theta} \langle S^\theta-P^\theta, A \rangle^{\alphabeta} \right].
    \end{align}

    \textbf{$(\theta,\alpha,\beta)$-AIM:} 
    Putting $g^{\biparamAIM}$ into \cref{eq:rmlr_pm_pt_simplified}, we have
    \begin{align}
        p(y=k \mid S \in \calM) 
        &\propto \exp \left[ \frac{1}{\theta^2}
        g^{\biparamAIM}_{\phi(P)} ( P^{\frac{\theta}{2}}\log(P^{-\frac{\theta}{2}} S^{\theta} P^{-\frac{\theta}{2}})P^{\frac{\theta}{2}},P^{\frac{\theta}{2}} \theta A P^{\frac{\theta}{2}} ) \right],\\
        &= \exp \left[ \frac{1}{\theta} \langle \log(P^{-\frac{\theta}{2}} S^\theta P^{-\frac{\theta}{2}}), A \rangle^{\alphabeta} \right].
    \end{align}    

    \textbf{$2\theta$-BWM}:   
    We first simplify \cref{eq:rmlr_pm_lt} under the cases of SPD manifolds and then proceed to focus on the case of $g=g^{\BWM}$.
    Denote $\phi: \{\spd{n},\tilde{g}, \tilde{\odot}\} \rightarrow \{\spd{n},g, \odot \}$, where the Lie group operation $\odot$ \citep{thanwerdas2022theoretically} is defined as
   \begin{equation}
        S_1 \odot S_2 = L_1 S_2 L_1^T, \forall S_1,S_2\in \spd{n}, \text{ with } L_1=\chol(S_1).
    \end{equation}
    Note that $I$ is the identity element of $\{\spd{n}, \odot\}$, and for any $S \in \spd{n}$, the differential map of the left translation $L_S$ under $\odot$ is
    \begin{equation}
        L_{S*,Q}(V)=LVL^\top, \forall Q \in \spd{n}, \forall V \in T_Q\spd{n}, \text{ with } L=\chol(S).
    \end{equation}

    For the induced Lie group $\{\spd{n},\tilde{\odot} \}$, the left translation $\tilde{L}_{P \tilde{\odot} I_{\tilde{\odot}}^{-1}}$ under $\tilde{\odot}$ is
    \begin{align}
        \tilde{L}_{P \tilde{\odot} I_{\tilde{\odot}}^{-1}}
        &= \phi^{-1} \circ L_{\phi(P)\odot \phi(I)^{-1}_{\odot}} \circ \phi,\\
        &= \phi^{-1} \circ L_{P^{2\theta}} \circ \phi. \quad (\phi(P)\odot \phi(I)^{-1}_{\odot}=P^{2\theta})
    \end{align}
    The associated differential at $I$ is
    \begin{align}
        \tilde{L}_{P \tilde{\odot} I_{\tilde{\odot}}^{-1} *,I} (A)
        &= \phi^{-1}_{*,\phi(P)} \circ L_{P^{2\theta} *,\phi(I)} \circ \phi_{*,I} (A),\\
        &= 2\theta \phi^{-1}_{*,\phi(P)} (\bar{L} A \bar{L}^\top),
    \end{align}
    where $\bar{L}=\chol(P^{2\theta})$. Then the SPD MLRs under $\tilde{g}$ and $\tilde{\odot}$ by left translation is
    \begin{equation}
        p(y=k \mid S \in \calM) 
        = \exp \left[ 2\theta g_{\phi(P)}\left(\rielog_{\phi(P)} \phi(S), \bar{L} A \bar{L}^{\top} \right) \right],
    \end{equation}

    Setting $g=g^{\BWM}$ (We omit the scaling factor.), we obtain the SPD MLR under $2\theta$-BWM:
    \begin{align}
        p(y=k \mid S \in \calM) 
        & = \exp \left [ 2\theta \cdot \frac{1}{4\theta^2}  g_{\phi(P)}^{\BWM} \left(\rielog_{\phi(P)}^{\BWM}\phi(S), \bar{L} A \bar{L}^{\top} \right) \right],\\
        & = \exp \left[ \frac{1}{4\theta}\langle (P^{2\theta}S^{2\theta})^{\frac{1}{2}} + (S^{2\theta}P^{2\theta})^{\frac{1}{2}} -2P^{2\theta}, \calL_{P^{2\theta}}(\bar{L} A \bar{L}^\top) \rangle \right].
    \end{align}
\end{proof}

\subsection{Proof of \cref{lem:equi_lie_mlr}}
\linkofproof{lem:equi_lie_mlr}

\begin{proof}[Proof of \cref{lem:equi_lie_mlr}]
    During this proof, we use the ambient representation of tangent vectors.
    We only need to prove the following:
    \begin{equation}
        \pt{Q}{P} = L_{PQ^{-1} *,Q}, \forall P,Q \in \so{n}.
    \end{equation}
     
    Given rotation matrices $P, Q$ and a tangent vector $H \in T _Q\so{n}$, the parallel transportation \citep[Tab. 1]{boumal2011discrete} is 
    \begin{equation}
    \pt{Q}{P} (H) = P Q ^\top H = P Q^{-1} H.
    \end{equation}
    On the other hand, given a curve $c(t)$ over $\so{n}$, satisfying $c(0)=Q$ and $c'(0)=H$, the differential of the left translation $L _{P Q ^{-1}}$ at $Q$ is
    \begin{equation}
    L _{P Q ^{-1} *,Q} (H) = \left. \frac{d PQ ^{-1}c(t)}{dt} \right| _{t=0} =P Q ^{-1} H,
    \end{equation}
    which concludes the proof.
\end{proof}

\subsection{Proof of \cref{thm:lie_mlr}}
\linkofproof{thm:lie_mlr}

\begin{proof}[Proof of \cref{thm:lie_mlr}]
    \cref{lem:equi_lie_mlr} indicates we can use either parallel transportation or group translation.
    Putting the associated expressions from \cref{tab:riem_rotation} into \cref{eq:rmlr_final} + \cref{eq:A_by_pt}, one can directly obtain the results.
\end{proof}

\newpage
\section*{NeurIPS Paper Checklist}

\begin{enumerate}

\item {\bf Claims}
    \item[] Question: Do the main claims made in the abstract and introduction accurately reflect the paper's contributions and scope?
    \item[] Answer: \answerYes{} % Replace by \answerYes{}, \answerNo{}, or \answerNA{}.
    \item[] Justification: Our abstract and introduction (\cref{sec:intro}) accurately reflect the paper's theoretical and empirical contributions.

\item {\bf Limitations}
    \item[] Question: Does the paper discuss the limitations of the work performed by the authors?
    \item[] Answer: \answerYes{} % Replace by \answerYes{}, \answerNo{}, or \answerNA{}.
    \item[] Justification: We discuss the limitations specifically in \cref{app:sec:limitation_future_work}.

\item {\bf Theory Assumptions and Proofs}
    \item[] Question: For each theoretical result, does the paper provide the full set of assumptions and a complete (and correct) proof?
    \item[] Answer: \answerYes{} % Replace by \answerYes{}, \answerNo{}, or \answerNA{}.
    \item[] Justification: Assumptions are clearly claimed in each theorem, and all the proofs are presented in \cref{app:proof}.

    \item {\bf Experimental Result Reproducibility}
    \item[] Question: Does the paper fully disclose all the information needed to reproduce the main experimental results of the paper to the extent that it affects the main claims and/or conclusions of the paper (regardless of whether the code and data are provided or not)?
    \item[] Answer: \answerYes{} % Replace by \answerYes{}, \answerNo{}, or \answerNA{}.
    \item[] Justification: Implementation details are discussed in \cref{app:sec:exp_details}.

\item {\bf Open access to data and code}
    \item[] Question: Does the paper provide open access to the data and code, with sufficient instructions to faithfully reproduce the main experimental results, as described in supplemental material?
    \item[] Answer: \answerYes{} % Replace by \answerYes{}, \answerNo{}, or \answerNA{}.
    \item[] Justification: All datasets (\cref{app:subsubsec:datasets_spd,app:subsub:datasets_son}) are publicly available. The code will be released after the review.

\item {\bf Experimental Setting/Details}
    \item[] Question: Does the paper specify all the training and test details (e.g., data splits, hyperparameters, how they were chosen, type of optimizer, etc.) necessary to understand the results?
    \item[] Answer: \answerYes{} % Replace by \answerYes{}, \answerNo{}, or \answerNA{}.
    \item[] Justification: In \cref{app:sec:exp_details}, we present the experimental details for reproducing the results. 

\item {\bf Experiment Statistical Significance}
    \item[] Question: Does the paper report error bars suitably and correctly defined or other appropriate information about the statistical significance of the experiments?
    \item[] Answer: \answerYes{} % Replace by \answerYes{}, \answerNo{}, or \answerNA{}.
    \item[] Justification: Mean, STD, and max of K-fold results are presented in \cref{sec:experiments}.

\item {\bf Experiments Compute Resources}
    \item[] Question: For each experiment, does the paper provide sufficient information on the computer resources (type of compute workers, memory, time of execution) needed to reproduce the experiments?
    \item[] Answer: \answerYes{} % Replace by \answerYes{}, \answerNo{}, or \answerNA{}.
    \item[] Justification: Hardware is mentioned in \cref{app:subsec:hardware}.
    
\item {\bf Code Of Ethics}
    \item[] Question: Does the research conducted in the paper conform, in every respect, with the NeurIPS Code of Ethics \url{https://neurips.cc/public/EthicsGuidelines}?
    \item[] Answer: \answerYes{} % Replace by \answerYes{}, \answerNo{}, or \answerNA{}.
    \item[] Justification: No ethic issue.

\item {\bf Broader Impacts}
    \item[] Question: Does the paper discuss both potential positive societal impacts and negative societal impacts of the work performed?
    \item[] Answer: \answerNA{} % Replace by \answerYes{}, \answerNo{}, or \answerNA{}.
    \item[] Justification: No societal impact.
    
\item {\bf Safeguards}
    \item[] Question: Does the paper describe safeguards that have been put in place for responsible release of data or models that have a high risk for misuse (e.g., pretrained language models, image generators, or scraped datasets)?
    \item[] Answer: \answerNA{} % Replace by \answerYes{}, \answerNo{}, or \answerNA{}.
    \item[] Justification: No such risks.

\item {\bf Licenses for existing assets}
    \item[] Question: Are the creators or original owners of assets (e.g., code, data, models), used in the paper, properly credited and are the license and terms of use explicitly mentioned and properly respected?
    \item[] Answer: \answerYes{} % Replace by \answerYes{}, \answerNo{}, or \answerNA{}.
    \item[] Justification: Original papers and datasets have been cited.

\item {\bf New Assets}
    \item[] Question: Are new assets introduced in the paper well documented and is the documentation provided alongside the assets?
    \item[] Answer: \answerNA{} % Replace by \answerYes{}, \answerNo{}, or \answerNA{}.
    \item[] Justification: Code will be released after the review.

\item {\bf Crowdsourcing and Research with Human Subjects}
    \item[] Question: For crowdsourcing experiments and research with human subjects, does the paper include the full text of instructions given to participants and screenshots, if applicable, as well as details about compensation (if any)? 
    \item[] Answer: \answerNA{} % Replace by \answerYes{}, \answerNo{}, or \answerNA{}.
    \item[] Justification: This paper does not involve crowdsourcing nor research with human subjects.

\item {\bf Institutional Review Board (IRB) Approvals or Equivalent for Research with Human Subjects}
    \item[] Question: Does the paper describe potential risks incurred by study participants, whether such risks were disclosed to the subjects, and whether Institutional Review Board (IRB) approvals (or an equivalent approval/review based on the requirements of your country or institution) were obtained?
    \item[] Answer: \answerNA{} % Replace by \answerYes{}, \answerNo{}, or \answerNA{}.
    \item[] Justification: No human subjects.

\end{enumerate}

\end{document}